\documentclass[11pt]{article}

\usepackage[preprint]{acl}

\usepackage{times}
\usepackage{latexsym}
\usepackage{tabularx}

\usepackage[T1]{fontenc}

\usepackage[utf8]{inputenc}

\usepackage{microtype}

\usepackage{inconsolata}

\usepackage{graphicx}
\usepackage{multirow}
\usepackage{booktabs}
\usepackage{tabularx}
\usepackage{caption}
\usepackage{subcaption}
\usepackage{balance}
\usepackage{amsmath,amssymb,amsfonts}%
\usepackage{amsthm}%

\title{Revisiting Prompt Sensitivity in Large Language Models for Text Classification: The Role of Prompt Underspecification}

\author{Branislav Pecher$^\dagger$, Michal Spiegel$^\dagger$$^\clubsuit$, Robert Belanec$^{\spadesuit}$$^\dagger$, Jan Cegin$^\dagger$ \\
$^\dagger$ Kempelen Institute of Intelligent Technologies, Bratislava, Slovakia\\
$^{\spadesuit}$ Faculty of Information Technology, Brno University of Technology, Brno, Czechia \\
$^\clubsuit$ Faculty of Informatics, Masaryk University \\
\texttt{\{name.surname\}}@kinit.sk\\ }

\begin{document}
\maketitle

\begin{abstract}
Large language models (LLMs) are widely used as zero-shot and few-shot classifiers, where task behaviour is largely controlled through prompting. A growing number of works have observed that LLMs are sensitive to prompt variations, with small changes leading to large variations in performance. However, in many cases, the investigation of sensitivity is performed using underspecified prompts that provide minimal task instructions, weakly constrain the model's output space or provide no definition of label space. In this work, we argue that a significant portion of the observed prompt sensitivity can be attributed to prompt underspecification. We systematically study and compare the sensitivity of underspecified prompts, without instruction and/or label space definition, with prompts that provide clear instructions and label choices. Utilising performance analysis, label token probability analysis, and linear probing, we find that underspecified prompts exhibit higher performance variance and lower probability values for relevant label tokens, while instruction-prompts suffer less from such problems. Overall, our findings highlight the need for more rigour when investigating and mitigating prompt sensitivity, with the need for more detailed analysis of internal representations besides looking at final layer outputs.
\end{abstract}

\section{Introduction}



\begin{figure}[tbh]
    \centering
    \includegraphics[width=0.99\linewidth]{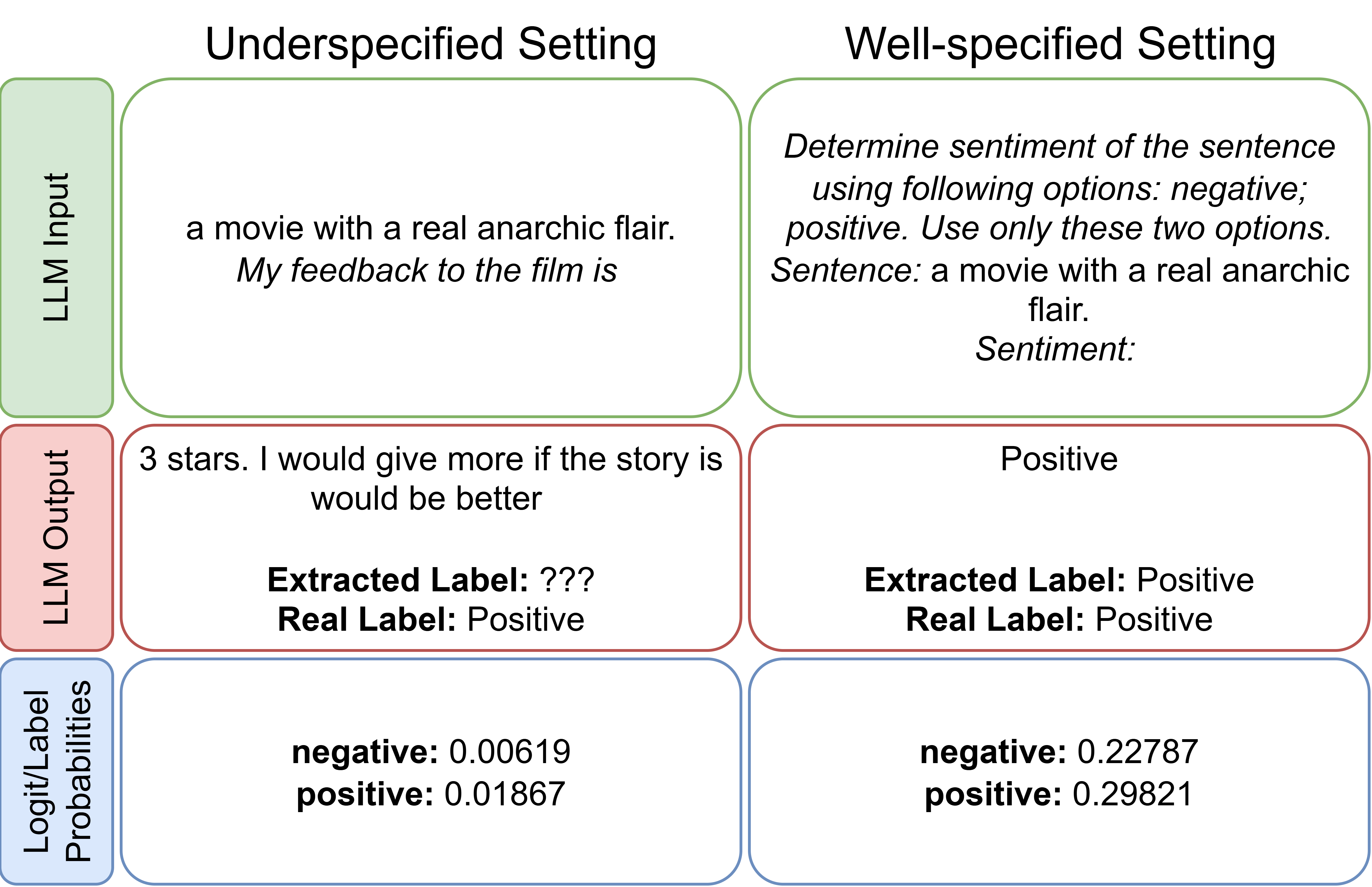}
    \caption{Using underspecified prompts leads lower label token probabilities and issues with label extraction.}
    \label{fig:introduction}
\end{figure}

Large language models (LLMs) have demonstrated strong performance across a wide range of tasks without the need for task-specific fine-tuning~\citep{brown2020language}. In particular, they are frequently used for text classification in zero-shot or few-shot settings, where the task is specified entirely through natural language prompts~\citep{liu2021pre}. Although it offers flexibility and reduced need for labelled data, much of the task specification is shifted to prompt design. As a result, prompt formulation has become an important factor affecting the model behaviour and performance~\citep{alzahrani2024benchmarks, sclar2024quantifying}.

A large number of existing works have observed that LLMs are highly sensitive to the phrasing of the prompt~\citep{sclar2024quantifying, habba-etal-2025-dove}. \textbf{In benchmarks or model comparisons, small changes in the prompt can lead to completely different rankings, changing what is considered state-of-the-art}~\citep{alzahrani2024benchmarks, habba-etal-2025-dove, voronov-etal-2024-mind}. Such behaviour has raised concerns about the robustness and reliability of prompt-based classification~\citep{zong2024fool}. To address these concerns, multiple mitigation strategies were proposed, including calibration~\citep{zhao2021calibrate, zhao-etal-2024-correcting, fei-etal-2023-mitigating}, prompt ensembling~\citep{jiang2023calibrating, voronov-etal-2024-mind}, or modifications to prompting strategies~\citep{zhou-etal-2022-prompt, chakraborty-etal-2023-zero}. 

However, the prompts used across these studies are varied. In case of typical text classification datasets, many of these studies rely on the prompts introduced by the original GPT-3 paper~\citep{brown2020language}. They often provide minimal task descriptions, instructions, possible label constraints, or no clear label space definition, essentially making them underspecified. For multi-choice question answering datasets (MCQA), the label space is implicitly defined (through the options), but often only minimal task descriptions are provided. As illustrated in Figure~\ref{fig:introduction}, underspecified prompts often cause LLMs to generate outputs that do not adhere to the expected format or include the expected label keywords. Researchers instead use the probability values of the final token to compare the probabilities of each class. However, as we show, these label probabilities are often extremely small due to the underspecification and the decisions are essentially made based on randomly distributed values. On the other hand, utilising well-specified prompts (i.e., instruction prompt formats) that provide the task instruction, possible label values and clear label space defition proved to solve most of these issues.




In this paper, we argue that the use of underspecified prompts is a significant contributor to the observed LLM prompt sensitivity. The models themselves have the capability of completing the task internally, regardless of the prompt used, but "fail" when outputting the results due to underspecification. To confirm this, we build on existing works and systematically compare the underspecified (also referred to as minimal) and well-specified (instruction) prompt formats across 3 different levels of specification: 1) providing minimal task description and no label space definition (underspecified prompts representative for text classification); 2) minimal task description with implicit label space definition (underspecified prompts representative for MCQA); and 3) well-specified instruction prompts with clear instructions and label space definition. We conduct the analysis across 3 text classification and 4 MCQA tasks and 10 LLMs, using: 1) empirical evaluation of accuracy and variance across 10 prompt variants per each setting; 2) analysis of token probabilities associated with the class labels, examining their values and distribution; and 3) mechanistic interpretability techniques, training linear probes on intermediate model activation to study how prompt (under)specification affects the internal representations. In addition, we investigate the impact of different strategies and simple modifications for mitigating sensitivity. Our contributions and findings are as follows
:
\begin{itemize}
    \item We analyse and demonstrate that prompt underspecification is a significant contribution of the prompt sensitivity, with instruction prompt formats yielding higher performance, lower variance, and more favourable token probability distribution compared to minimal prompts.
    \item Demonstrating different tools that can be used for rigorous investigation and mitigation of prompt sensitivity, including: 1) label token probability analysis, showing the probabilities are a good indicator of prompt quality, with positive correlation to accuracy as high as $0.757$; and 2) linear probe analysis, showing the internal representations of LLMs themselves are affected by underspecification, leading to lower label decodability and robustness.
    \item Showing that in-context learning can address the majority of the sensitivity stemming from prompt underspecification as effectively as calibration, without needing to access model internals.
\end{itemize}


\section{Related Work}

The problem of prompt sensitivity in LLMs has already drawn a lot of attention~\citep{zong2024fool, gonen-etal-2023-demystifying}. Some studies look at how using rephrased, but semantically similar prompts affects the models, finding significant variation in performance~\citep{mizrahi-etal-2024-state, errica-etal-2025-wrong}. Even small changes, such as using different separator, spacing or cases (e.g., upper/lowercase), can lead to significant changes~\citep{salinas-morstatter-2024-butterfly, sclar2024quantifying}, even changing what is considered to be state-of-the-art~\citep{voronov-etal-2024-mind, habba-etal-2025-dove}. Similarly, even changing what labels are used for classification or the order of options leads to significant changes~\citep{molfese-etal-2025-right, alzahrani2024benchmarks}. Other studies look at how different prompt characteristics, such as emotional tone, roleplaying or output requirements affect the performance~\citep{zhuo-etal-2024-prosa}. Further studies look at how the structure of the prompt, such as using chain-of-though, affects the peformance~\citep{lu-etal-2024-prompts}). On the other hand, \citet{molfese-etal-2025-right} suggest that the problem stems from the extraction methods used in evaluation that tend to underestimate the LLM abilities.

To address this problem, various mitigation strategies are employed. Some approaches rely on the observation that LLMs tend to prefer specific tokens and give them higher probabilities~\citep{zhao-etal-2024-correcting}. To address this, the skew in the label probabilities is first determined using empty prompts~\citep{zhao2021calibrate}, random words from a text corpus~\citep{reif-schwartz-2024-beyond}, batches of samples~\citep{zhou2024batch}, linear probes~\citep{abbas2024enhancing}, ICL samples~\citep{reif-schwartz-2024-beyond} or average probabilities from unlabelled samples~\citep{zhao-etal-2024-correcting} and then is calibrated. Instead of calibration, \citet{qian-etal-2025-beyond} suggests appending a large number of empty (e.g., "UNK") tokens to the prompts and using an ensemble of the label probabilities across these tokens to normalise the distribution. Further works suggest using an ensemble of multiple prompts to deal with the sensitivity, either only in the inference stage~\citep{voronov-etal-2024-mind, jiang2023calibrating} or in training~\citep{zhou-etal-2022-prompt}. Another set of approaches suggests automatically and iteratively generating and selecting good-performing prompts~\citep{errica-etal-2025-wrong, chakraborty-etal-2023-zero}. 

Many of these works, especially those focused on typical text classification, utilise underspecified prompts that do not contain any task description or instructions (or only implicit label space definition in case of MCQA), which may contribute to the observed sensitivity~\citep{zhao2021calibrate, fei-etal-2023-mitigating, zhou2024batch, abbas2024enhancing, voronov-etal-2024-mind, qian-etal-2025-beyond}. However, recent works that use well-specified prompts~\citep{salinas-morstatter-2024-butterfly, mizrahi-etal-2024-state, alzahrani2024benchmarks, errica-etal-2025-wrong, habba-etal-2025-dove, sclar2024quantifying} also observe large effects of prompt sensitivity, albeit often from larger changes. In this paper, we build and complement these works (especially~\citet{habba-etal-2025-dove, sclar2024quantifying, zhuo-etal-2024-prosa}) and take a closer look at how the different levels of prompt (under)specification, defined by clear instruction and label space mapping, affect the observed prompt sensitivity in LLMs, while also investigating the impact of existing mitigation strategies on various prompting regimes.

\section{Analysing Prompt Sensitivity}

The goal in this section is to analyse the sensitivity of large language models to prompt variations and the impact of prompt underspecification on this sensitivity, through the lens of the most commonly used metric -- the final classification performance of the LLM measured through accuracy. To explore the impact of underspecification, we utilise two separate prompt formats. The first, which we call \textbf{minimal} or underspecified, is inspired by the original GPT-3 paper~\citep{brown2020language}, where we define two levels of underspecification based on the used task: 1) those containing no instruction or label space definition for text classification; and 2) those containing no instruction and only implicitly defined label space for multi-choice QA. As the variations of these prompts are commonly used when analysing prompt sensitivity, we draw these prompts from existing works~\citep{zhao2021calibrate, qian-etal-2025-beyond, sclar2024quantifying}. The second, \textbf{instruction} prompt format, provides the LLM with specific instructions on what task to perform (e.g., "Determine sentiment") and the possible labels that can be assigned (e.g., negative or positive for sentiment classification). These prompts are designed based on those used in related works that either directly use them or provide recommendations on how such prompts should be constructed~\citep{errica-etal-2025-wrong, alzahrani2024benchmarks, sun2023pushing}. We use 10 prompts of each kind to determine the sensitivity to prompt variations. In cases where we are unable to identify 10 sufficiently different prompts for the given dataset (which mostly occurs with the \textbf{instruction} prompts), we use GPT-4o to generate new ones using a subset of the available prompts as examples. We manually check these prompts for biases and disregard the ones that are too out of distribution or too similar to the ones used as examples. The prompt formats for each dataset, as well as for generating new ones, are provided in Appendix~\ref{app:experimental-details}.

Furthermore, we investigate the impact of various strategies for mitigating sensitivity to prompt variations, drawing on related work. As a baseline, we use the \textit{Base} setting, where the LLM and prompt are used as is. As the first mitigation strategy, we employ a recently proposed strategy (denoted as \textit{UNK} from now on)~\citep{qian-etal-2025-beyond} that addresses underspecification by appending a larger number of "<unk>" tokens to the prompt. Second, we use the popular \textit{calibration}~\citep{zhao2021calibrate, fei-etal-2023-mitigating, zhou2024batch, reif-schwartz-2024-beyond} mitigation strategy that first determines the skew in the LLM token probabilities using neutral prompts and then calibrates them to guarantee the probabilities for the labels are uniformly distributed (e.g., 0.5 probability for both negative and positive classes in the neutral prompt for sentiment classification). We explore multiple possibilities for designing the neutral prompt based on related work and report only the best-performing one (determined as mean across all models and datasets) -- using "empty" prompt (e.g., providing an empty string as the sample for classification) as used by~\citet{zhao2021calibrate}. 

As we hypothesise that most strategies were designed to address underspecification in prompts indirectly, our goal is to determine whether they provide meaningful impact even when the problem of underspecification is addressed using simple LLM modifications and prompting strategies. Therefore, we compare the impact of these strategies with simple LLM modifications. First, we utilise \textit{in-context learning}, where we modify the prompt by providing 2 samples per class as part of the prompt. Second, instead of using the base version of the model, we use the \textit{instruction-tuned} variants of the models. Third, we utilise the (meta-)\textit{tags} that delimitate the different parts of the prompt (e.g., system part, user part, and the model part) and use the chat-template functionality provided by the tokenisers. Finally, we also apply different strategies for the instruction-tuned variants of the models to determine whether such a combination of simple modification and mitigation strategies provides any performance benefit.

Following the recent observations that traditional evaluation strategies often underestimate LLM capabilities~\citep{molfese-etal-2025-right}, we use two separate evaluation strategies: 1) \textit{label token probability evaluation}; and 2) \textit{generation evaluation}. In both cases, we report the mean accuracy and its standard deviation across the 10 prompt variations.

\paragraph{Label Token Probability Evaluation.} In the label token probability evaluation strategy, the answer or class assigned by the LLM is obtained by passing the prompt to the model and analysing the probability distribution after its last token. The softmax function is applied to the model's logits from the last token, resulting in a value between 0 and 1 for each token in the model's vocabulary. The probabilities for the tokens corresponding to each label (e.g., `positive` or `negative` for sentiment classification) are compared, and the token with the highest probability is assigned as the class. Although easy to use, this method ignores the magnitude of the probability values, which may lead to effectively random classification decisions driven by low probabilities.

\paragraph{Generation Evaluation.} In the generation evaluation strategy, the answer or class is extracted from the free text generated by the LLM after passing the prompt to the model. The text is automatically checked (using string matching) for the occurrence of the possible labels. If only a single label is found in the answer, it is assigned; otherwise (e.g., more than one label appears or no label appears as part of the text), no label is assigned. The dataset-specific details are included in Appendix~\ref{app:experimental-details}.

\paragraph{Models.} For evaluation, we choose models of various sizes from the LLaMA family of models~\citep{touvron2023llama, grattafiori2024llama} -- \textbf{LLaMA-3.1-8B}, \textbf{LLaMA-3.2-3B}, \textbf{LLaMA-2-13B} -- and \textbf{Qwen-3-4B}~\cite{qwen3technicalreport} and \textbf{Gemma-3-4B}~\cite{Kamath2025Gemma3T} in both the base and the instruction-tuned variants. Each model is run in a deterministic fashion, setting the temperature to 0, using no sampling and generating a maximum of 50 tokens.

\paragraph{Datasets.} The experiments are conducted on 3 text classification datasets and 4 multi-choice question answering datasets, each comprising different tasks with varying complexity. The text classification datasets are \textbf{SST2}~\citep{socher-etal-2013-recursive} for sentiment classification with 2 classes (negative, positive), \textbf{AG News}~\citep{zhang2015agnews} for news classification with 4 classes (world, sports, business, science and technology), and \textbf{ISEAR}~\citep{scherer1994evidence-isear} for emotion detection with 7 classes (shame, joy, sadness, disgust, anger, fear, guilt). The multi-choice QA datasets are \textbf{MMLU}~\citep{hendrycks2021measuring}, \textbf{ARC}~\citep{allenai:arc} (including both Challenging and Easy splits) with 4 classes (A, B, C, D) and \textbf{Commonsense QA}~\citep{talmor-etal-2019-commonsenseqa} with 5 classes (A, B, C, D, E). Following the common practice, we use the provided test or validation splits for evaluation.

\subsection{Results}
\label{sec:emprirical-eval}

\begin{figure*}[tbh]
    \centering
    \includegraphics[width=.9\textwidth]{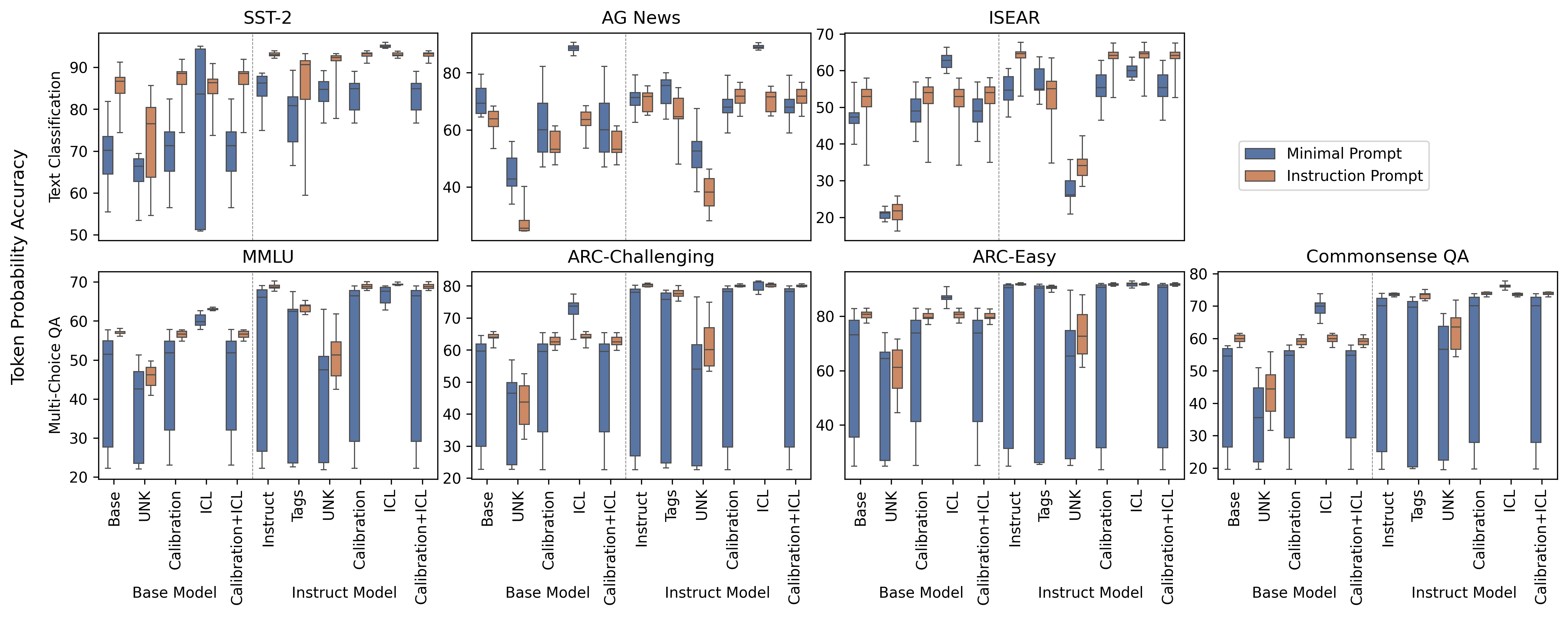}
    \caption{Sensitivity of LLaMA-3.1-8B model to prompt variations measured as token probability accuracy and its deviation across different prompts. The problem of underspecification can be solved using a combination of instruction prompts and in-context learning.}
    \label{fig:main-results-llama3}
\end{figure*}

Our focus is on answering 2 main questions: \\
1) \textit{is prompt sensitivity driven by underspecification, and can it be effectively mitigated with instruction-based prompting}; and \\
2) \textit{what is the impact of simple LLM prompt modifications, sensitivity mitigation strategies, and their combination on the prompt sensitivity}. To answer these questions, Figure~\ref{fig:main-results-llama3} provides accuracy and standard deviation across different datasets, mitigation strategies and prompt formats for the LLaMA-3.1-8B model and label token probability evaluation. The results for all models and evaluation strategies are included in Appendix~\ref{app:main-results-further-models}.

\textbf{Although the instruction prompts often reduce the impact of underspecification, they are not enough to deal with all sensitivity by themselves.} Based on the label token probability evaluation, the instruction prompt formats often lead to markedly higher mean performance, and we observe significantly lower deviation or spread of the results. However, in some cases, the instruction formats underperform the minimal ones, such as on the AG News dataset for all models, or the text classification datasets (SST2, AG News, and ISEAR) on LLaMA-3.2-3B and LLaMA-2-13B models. In multiple cases, we not only observe lower performance, but also higher standard deviation, which indicates problems with specific prompts. As the observed performance is often lower than in the cases where instruction formats are consistently better, these results may indicate problems stemming from the datasets (e.g., information leak) or the models themselves (e.g., smaller or older versions of the models not working well with longer instructions). In addition, this is mostly a case when the base version of the models is used, which may further point to problems with the underlying models that is not optimised for instruction following. Looking at the generation performance evaluation (see Appendix~\ref{app:main-results-generation}) the benefit of instruction prompt formats is clearer. The minimal prompt format often fails to even outperform random chance. However, such a result can be expected, as these prompts are not designed to elicit a relevant response and the models in these cases simply try to complete the provided prompt (e.g., in the case of sentiment analysis, providing a review for the movie with neutral stance).

\textbf{The benefit of instruction prompt formats is higher on the multi-choice question answering datasets.} Even though these datasets already provide implicit label space definition (through the options) and thus provide more relevant prompts, using clearer and more explicit instruction still leads to large benefits. The benefit is most noticeable on the deviation of the results, which is significantly lower (e.g., the spread of performance is only on average $2-3\%$ for instruction prompts as compared to $20-30\%$ for minimal prompts). On the other hand, the average performance across the prompts might not always be markedly higher ($3\%$ on average; even lower for specific models).



\textbf{Simple modifications lead to strong mitigation of prompt sensitivity.} Just by switching to the instruction-tuned variant of the models, we observe average increases in performance of around $1-15\%$ for the minimal prompts and $6-21\%$ for the instruction prompts in both terms of label token probability and generation accuracy. The higher benefit for the instruction prompt formats might be due to better alignment of the prompt format with the model, better instruction following capabilities, but also due to possible information leak (e.g., models being possibly trained on the test sets of these datasets). However, as we are mainly interested in comparing minimal and instruction prompts on the same setup, such information leak does not necessarily cause problems.

\textbf{In-context learning provides the most consistent benefit for dealing with LLM prompt sensitivity and underspecification.} Across all models and datasets, in-context learning provides the highest performance increase and reduction in standard deviation, especially when combined with the instruction-tuned variants of the models. The benefit is observed for both the minimal and instruction prompt formats. Furthermore, for the minimal prompt formats, using in-context learning even leads to a significant increase in generation accuracy, even though it does not reach the performance of instruction formats. These results indicate that the prompt sensitivity is largely caused by underspecification, as the models are uncertain about what the task is and how it should be solved. \textbf{Calibration provides similar, although slightly smaller and less consistent improvements.} The differences in performance between calibration and in-context learning are as high as $+1.5\%$ (LLaMA-3.1-8B on SST2) but as low as $-8\%$ (LLaMA-3.1-8B on AG News). However, the combination of in-context learning and calibration leads only to minor further improvements, suggesting that both are targeting the same issues. On the other hand, using the \textit{UNK} mitigation strategy often leads to a significant degradation of performance, even resulting in models performing worse than random chance, as seen with the LLaMA-3.2-3B model. As such, \textbf{the strategy of using \textit{UNK} tokens is not suitable for dealing with prompt sensitivity.} Finally, using meta-tags and the chat template from tokenisers provides benefits only in terms of generation performance, while hurting the logit performance.

To summarise, we can answer the questions in this section as: 1) \textit{the problem of underspecification cannot be fixed solely by using instruction prompt formats, even though their use often helps}; and 2) \textit{the combination of using in-context learning and instruction-tuned model variants deals with most problems of underspecification and LLM prompt sensitivity, especially when combined with instruction prompt formats}. As such, \textbf{a large part of the observed prompt sensitivity is due to the use of underspecified prompts and problems in the experimental setup} (such as using models and prompts not suited for the problem at hand). Finally, \textbf{both clear instruction and label space definition are important when it comes to prompt sensitivity}, as observed by the benefit of instruction formats over minimal ones on the MCQA datasets that provide implicit label space definition.

\section{Label Token Probability Analysis}
\label{sec:logit-analysis}

In this section, our goal is to better understand the internal behaviour of LLMs using a metric beyond simple performance. Although the accuracy is a common measure, it does not provide additional information about the cause of prompt sensitivity. Therefore, motivated by the observed results from the previous section, where the instruction prompt formats and in-context learning lead to improvements in regards to underspecification, but not in all cases, we turn towards a more detailed analysis of the token probabilities from the last LLM layer -- as a common practice of sensitivity mitigation strategies is addressing the underlying probability distributions. Our main focus is on answering the following research question: 1) \textit{is there any connection between the observed performance, its variation across prompts, and the underlying token probabilities?}  For this analysis, we follow the steps as described in the \textit{label token probability evaluation}, but instead of considering only the label token with the highest probability, we collect the probabilities for all considered classes, as well as the top 20 tokens from the LLM vocabulary with the highest assigned probability. The distribution of label token probabilities, averaged over the individual classes and all prompts are presented in Figure~\ref{fig:token-probabilities-llama3} for LLaMA-3.1-8B model and in Appendix~\ref{app:probabilities-other-models}.

\begin{figure*}[tbh]
    \centering
    \includegraphics[width=.85\textwidth]{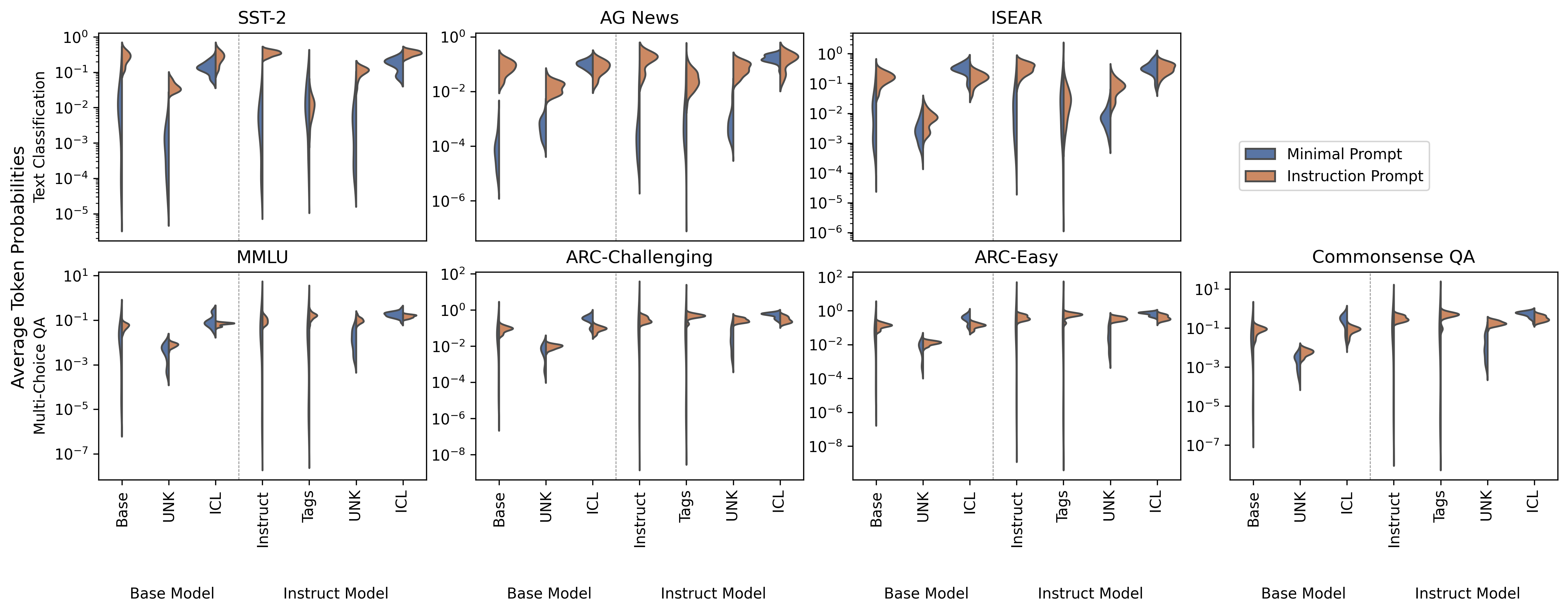}
    \caption{Token probability distribution across different prompts for all labels for LLaMA-3.1-8B model. The instruction prompt formats and in-context learning consistently lead to highest label token probability values. On the other hand, the probabilities for minimal prompt formats are significantly smaller with larger distribution.}
    \label{fig:token-probabilities-llama3}
\end{figure*}

\textbf{Issue of underspecification can be clearly observed in the label token probability distribution.} For the minimal prompt formats, the probabilities are distributed across large range of very small probability values (in the range of $10^{-5}$ to $10^{-3}$\footnote{The label tokens are almost always in the lower $50\%$ of the most probable tokens from the model dictionary, with the most probable tokens being completely irrelevant}. Such low probabilities and their distribution raise questions about the reliability of classification evaluations based on pure token probabilities, as decisions are made on more or less randomly distributed values and may lead to an overestimation of the LLM's capabilities. In addition, such decisions may also be a significant contributor to the prompt sensitivity. However, in specific cases, even when the token probability values are smaller, we can observe high performance, such as LLaMA-3.2-3B on the SST2 dataset. \textbf{For the instruction prompt formats, we often observe higher label token probabilities that are better distributed.} We observe a statistically significant difference between the distributions (p-value of $2.088e-110$ using Mann-Whitney U-test). Furthermore, in cases when the instruction prompt formats underperform, especially on the LLaMA-3.2-3B and LLaMA-2-13B models, or the AG News dataset, we can observe a drop in the probability values. Finally, \textbf{using the best performing sensitivity mitigation strategy, in-context learning, almost always results in high probability values.} However, for the instruction prompt formats, the changes in distribution when using in-context learning is not statistically significant (p-value of $0.589$ using Mann-Whitney U-test). These results indicate the importance of label token probability values to deal with underspecification and prompt sensitivity -- well-distributed values lead to higher performance and lower deviation in results.

\begin{figure}[tbh]
    \centering
    \includegraphics[width=.9\linewidth]{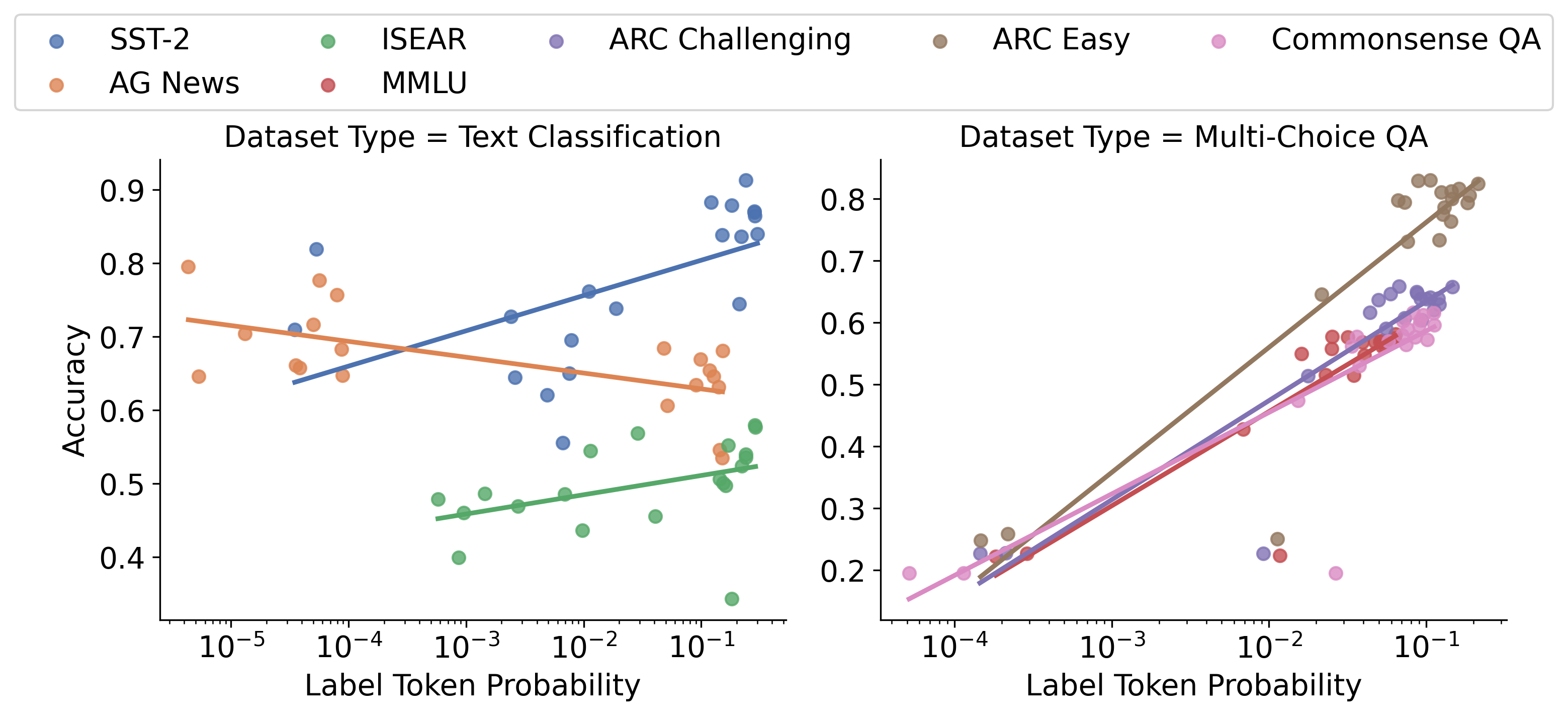}
    \caption{Correlation between label token probabilities and performance metric for prompts. The label token probabilities are a good indicator of prompt quality.}
    \label{fig:probability-accuracy-correlation}
\end{figure}

To provide further analysis, we look at specific prompts (instead of simple averages). For each prompts, we plot the correlation between the performance it achieves and the average label token probabilities for all classes in Figure~\ref{fig:probability-accuracy-correlation}. We can observe that \textbf{high label token probabilities are a good indicator of the prompt quality.} Irrespective of what prompt format is used, higher label token probabilities lead to higher performance. Similarly, we observe positive correlation between the token probabilities and accuracy -- calculated as Pearson correlation, the values are $0.757$, $0.532$, and $0.337$ for the LLaMA-3.1-8B, LLaMA-3.2-3B, and LLaMA-2-13B models, respectively. The lower correlation values for the LLaMA-3.2-3B and LLaMA-2-13B models are mainly due to them achieving higher performance even with lower probability values. \textbf{Even though higher performance can be achieved even with low label token probabilities and their worse distribution caused by underspecification, prompts that achieve higher label token probabilities always lead to better performance.}

In summary, we can conclude that \textit{using label token probability analysis can be used to identify high-quality prompts that suffer less from underspecification, and lead to better performance.}


\section{Linear Probe Analysis}
\label{sec:linear-probes}

\begin{figure*}[!tbh]
    \centering
    \includegraphics[width=.8\textwidth]{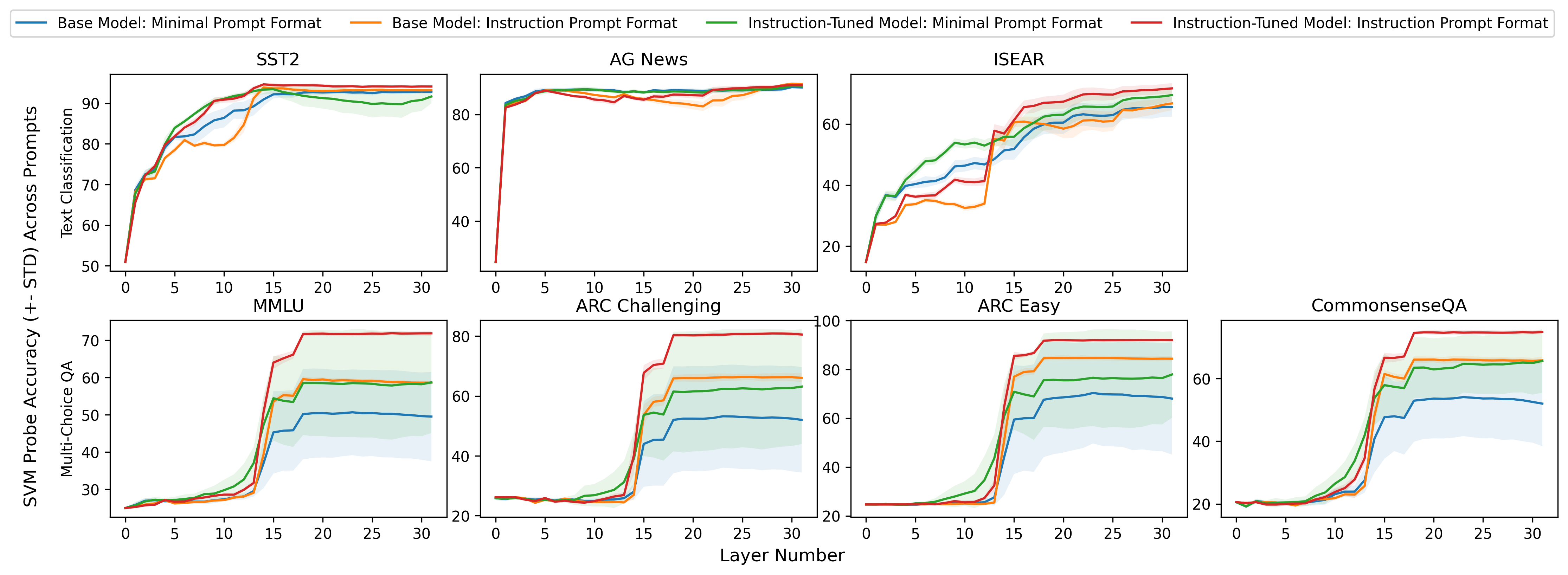}
    \caption{The mean accuracy of the SVM model over different minimal and instruction prompt formats for all layers of the LLaMA-3.1-8B model. Minimal prompts affect the internal representations and the model label decodability, especially for multi-choice QA datasets.}
    \label{fig:SVM-probe-llama3}
\end{figure*}

So far, the analysis has been done based on the final layer of the LLM. The goal of this section is to examine the behaviour of the LLM throughout the whole network and answer the question of how the individual layers of the LLM are affected by the underspecification and different prompt formats. We analyse whether the labels are decodable from the internal representations of the LLM, and how the underspecified prompts affect this ability. To perform this analysis, we employ linear probe analysis. Using the training set of each dataset, we create prompts for each observation, pass them to the LLM, collect the outputs of each layer and use these outputs to train a separate SVM classifier (with RBF kernel) for each layer. Afterwards, we repeat the same process for each test sample and evaluate the accuracy of the SVM model, on each layer, in predicting the final sample label. We perform this for all prompts, datasets, and models. The mean accuracy over the individual prompts for all layers is presented in Figure~\ref{fig:SVM-probe-llama3} for the LLaMA-3.1-8B model. In Appendix~\ref{app:linear-probe-analysis} we provide results for the remaining models, along with further details and explanations of why we use an SVM model instead of simple linear regression.

\textbf{Internal representations of the LLM and their ability for label decodability is affected by minimal prompts}, especially on more complex datasets. Focusing on the simpler task (classifying labels based on the outputs of individual layers), we can observe that starting from middle layers, the SVM model is able to achieve high performance that is only slightly lower than when using in-context learning. On the text classification datasets (SST2, AG News and ISEAR), the difference in performance between instruction and minimal prompts depends on the complexity of the dataset. On the simpler SST2 and AG News datasets, the difference between the different prompts formats, but also different model variants is quite small. On the other hand, on the more complex ISEAR dataset, we observe that using instruction prompt formats leads to higher SVM accuracy with lower deviation. On the other hand, there is a clear distinction between the formats on the multi-choice QA datasets. The minimal prompts format lead to performance that is on average lower by $10\%$, while also leading to markedly higher deviation in results. At the same time, the instruction prompt formats from the base version of the models often outperform the minimal prompt formats from the instruction-tuned variants of the models. Same results can be observed for the remaining models as well, while in all cases, the differences are statistically significant (p-value of $1e-2$ to $1e-15$). As such, we can conclude that the \textit{underspecification from prompts has an impact on the degradation of the internal processing of the models}. Therefore, linear probing can serve as an additional tool for understanding prompt sensitivity, especially on datasets with implicit label space definition (more complex MCQA datasets), in order to improve the task instructions.

\section{Conclusion}


This study identifies prompt underspecification as a primary driver of LLM prompt sensitivity. Our results demonstrate that instruction-based formats significantly improve performance and label token probability distribution compared to minimal prompts that lack clear task instructions and/or label space definition. While prompt underspecification significantly drives sensitivity by increasing variance and suppressing performance, these issues are effectively mitigated through instruction-based formats, in-context learning, instruction-tuning, and calibration. Through label token probability analysis, we have observed a strong correlation ($r = 0.757$) between token probabilities and prompt performance. Furthermore, linear probe analysis revealed that internal task representations are affected by underspecification as well, degrading performance even in the simpler label decodability task. Importantly, we show that in-context learning mitigates this sensitivity as effectively as calibration, without requiring model internal access. These findings suggest that a significant portion of observed LLM instability stems from underspecified prompts and ambiguous instructions rather than a fundamental inability to solve the task. We therefore advocate for greater methodological rigour, emphasising well-specified prompts to ensure more reliable assessments of model capabilities.


\clearpage

\section*{Limitations}

Due to the extent and nature of our analysis, which focuses more on detailed qualitative analysis rather than quantitative analysis, we perform experiments across 7 datasets of varying characteristics and complexity (binary sentiment classification, news classification, emotion detection and multi-choice question answering). In addition, due to the characteristics of some mitigation strategies, we focus only on 5 models (or 10 if the base and instruction-tuned versions of the same model are treated as separate models) of different sizes (3B, 4B, 8B and 13B parameters) from the LLaMA family, Qwen and Gemma. Running the analysis across a larger number of datasets and models could further fine-tune the findings, but would introduce further problems (e.g., finding specific "empty" tokens for different model families that could be used instead of the UNK tokens) and would significantly increase the required compute.

Furthermore, we focus only on two aspects of the underspecification, providing clear instructions and label space mapping. Looking at further possibilities for providing specified prompts, such as exact label descriptions, or other improvements, such as chain-of-thought prompts, could be interesting, but would significantly increase the size of experiments. In addition, we explore large changes in the prompts, where the whole prompts are reformulated, instead of exploring the impact of smaller changes (e.g., changing individual words or characters with no semantic meaning). However, as we focus on more detailed qualitative analysis we leave the detailed analysis of these scenarios using the tools shown in the paper as future work.

Evaluating the model performance based on the generated text is done using heuristic extraction using string matching and regex, which may underestimate the capabilities of the models in this regard as observed by prior works (e.g.,~\citep{molfese-etal-2025-right}). However, we opted for this setup after manually checking the generated outputs and evaluating the possible underspecification, finding its effects to be not as significant in this regard. This issue could be solved by using an LLM judge that would extract the meaning from the sentences, in a similar fashion to~\citet{molfese-etal-2025-right}, but it would introduce further compounding factors (e.g., the LLM could introduce further biases into the results). As such, we opted for the simpler setting.

Finally, we perform the analysis of prompt sensitivity across 10 prompts of different formats (both minimal and instruction prompt formats). Increasing the number further could potentially improve the analysis. However, using 20 prompts already goes beyond the common practice in the existing works. In addition, increasing the number would require us to extensively use an LLM to generate them, which could introduce further biases, as already for this paper we needed to perform extensive manual checks as the LLM often generated less than usual prompts. 

\section*{Ethical Considerations}
The experiments in this paper work with publicly available datasets, citing the original authors. We do not work with any personally identifiable information or offensive content and perform no crowdsourcing for further data annotation. In addition, we are not aware of any potential ethical harms or negative societal impacts of our work, apart from the ones related to the advancement of the field of machine learning. We follow the license terms for all the models we use. It is possible that the large language models we use contain biases and potentially offensive or harmful content. However, the original authors of these models reduce this bias as much as possible. Finally, as the highest impact of our study is the CO2 generated as part of our experiments, we publicly release the impact statement and the amount of compute used below.

\paragraph{Impact Statement: CO2 Emissions Related to Experiments} The experiments presented in this paper used significant computational resources as they require multiple evaluation runs of multiple models (to determine sensitivity over larger number of prompts), as well as using large language models that require a lot of computation even just for the inference. Overall, the experiments were conducted using a private infrastructure, which has a carbon efficiency of 0.432 kgCO$_2$eq/kWh (default value used as the actual efficiency of our HW instance was not measured). A cumulative of 1500 hours of computation was performed on hardware of type A100 PCIe 80GB (TDP of 250W). Total emissions are estimated to be 162 kgCO$_2$eq of which 0 percents were directly offset. This includes all of the inference with all models and training and evaluation of the linear probes, and the preliminary experiments. These estimations were conducted using the \href{https://mlco2.github.io/impact#compute}{MachineLearning Impact calculator} presented in \cite{lacoste2019quantifying}. Whenever possible, we tried to reduce the compute resources used as much as possible. For example, collecting the logit values at the same time as the empirical performance evaluation, or using calibration of the logit post-hoc (e.g., calculating the skew separately and applying it to the collected logit probabilities).

\bibliography{references}

\appendix

\section{Additional Results}

\subsection{Main Results: Additional Models}
\label{app:main-results-further-models}

\begin{figure*}[!th]
     \centering
     \begin{subfigure}[t]{1\linewidth}
        \centering
        \includegraphics[width=1\linewidth]{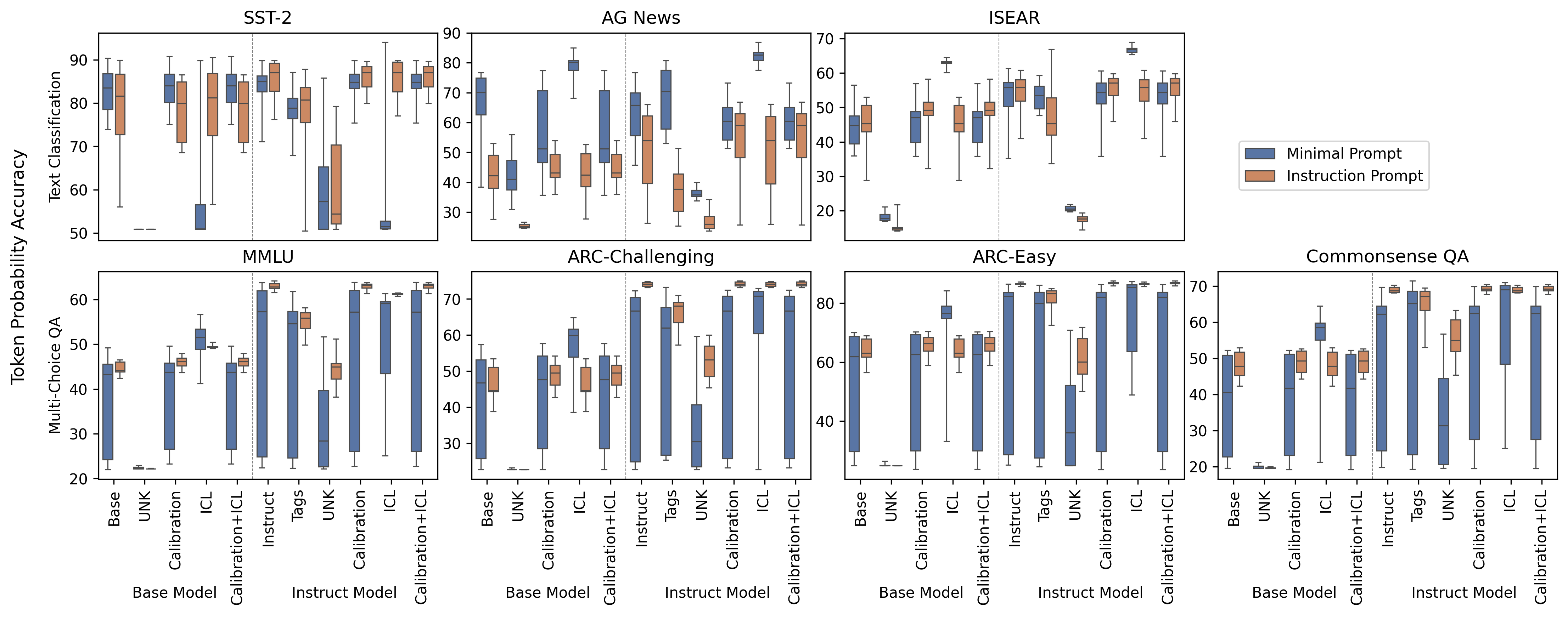}
        \subcaption{LLaMA-3.2-3B}
    \end{subfigure}
    \hfill
    \begin{subfigure}[t]{1\linewidth}
        \centering
        \includegraphics[width=1\linewidth]{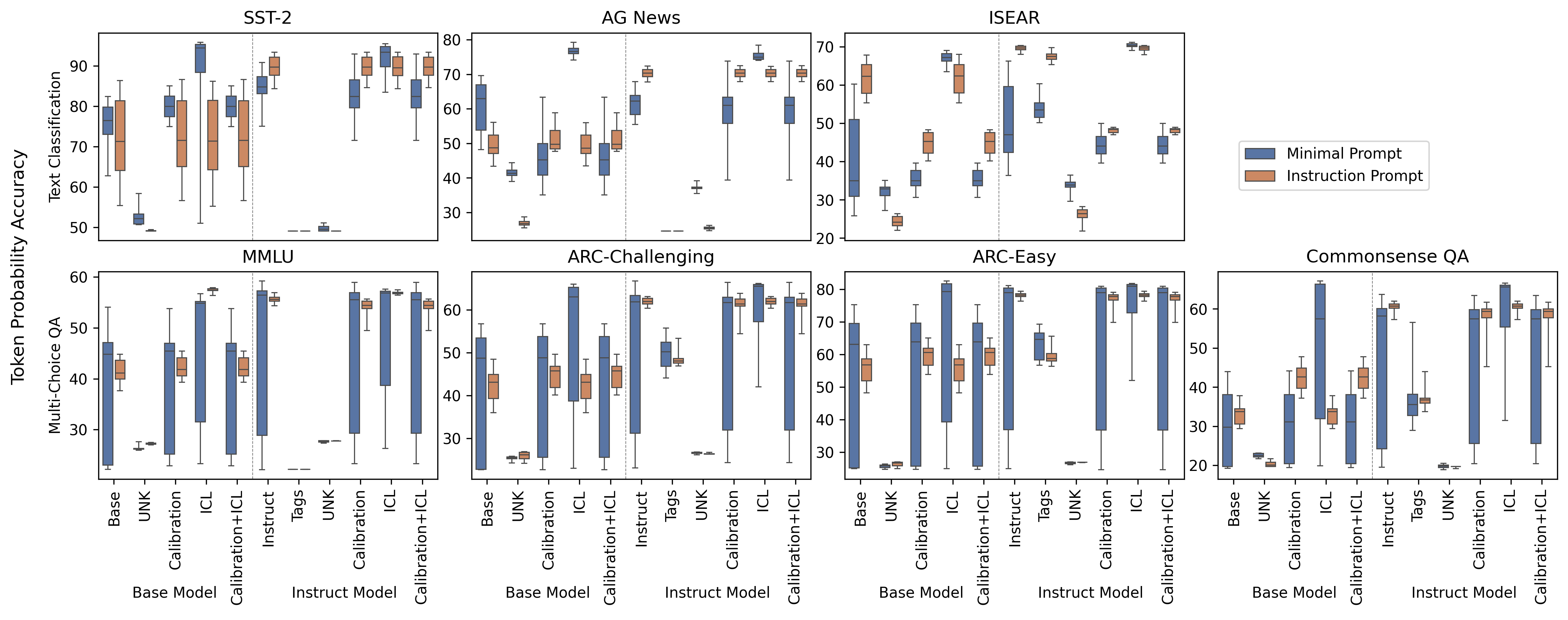}
        \subcaption{LLaMA-2-13B}
    \end{subfigure}
    
    \caption{Sensitivity of LLaMA-3.2-3B (a), LLaMA-2-13B (b) models to prompt variations measured as token probability accuracy and its deviation across different prompts. The problem of underspecification can be solved using a combination of instruction prompts and in-context learning.}
    \label{fig:main-results-further-models}
\end{figure*}

\begin{figure*}[!th]
     \centering
    \begin{subfigure}[t]{1\linewidth}
        \centering
        \includegraphics[width=1\linewidth]{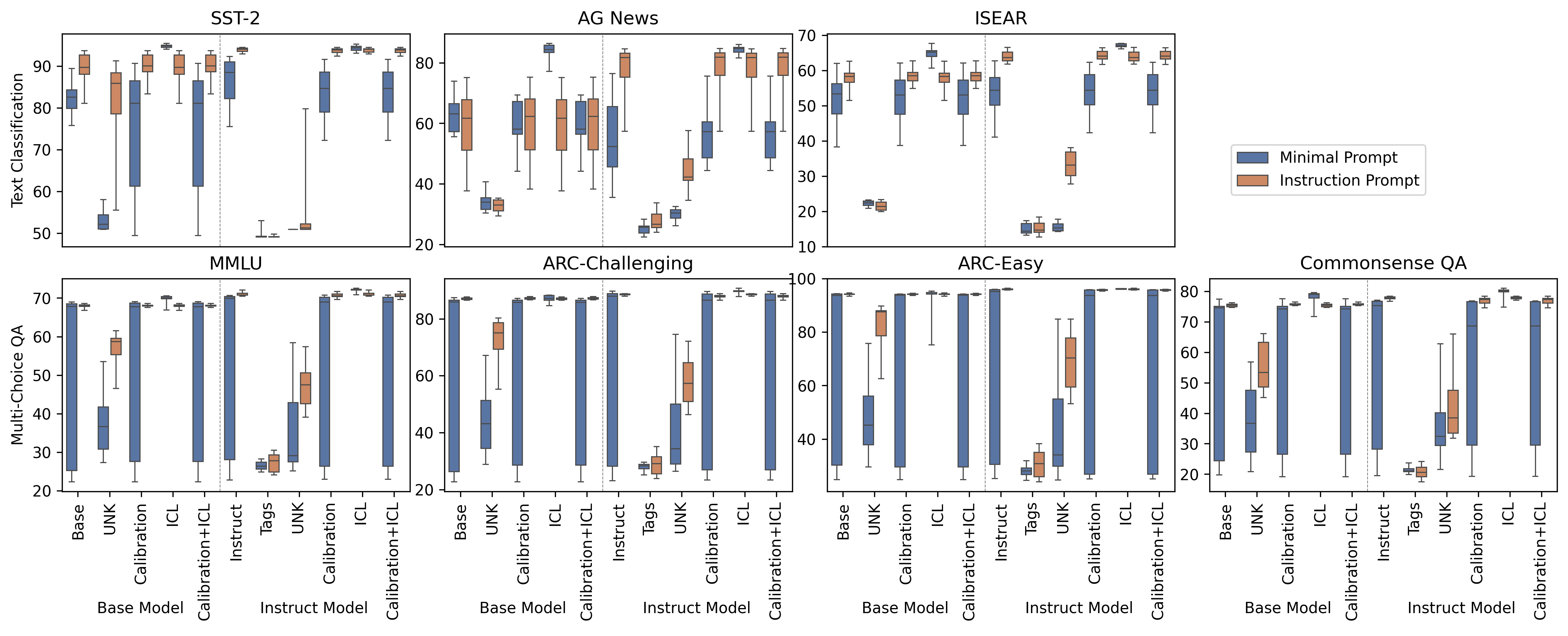}
        \subcaption{Qwen-3-4B}
    \end{subfigure}
    \hfill
    \begin{subfigure}[t]{1\linewidth}
        \centering
        \includegraphics[width=1\linewidth]{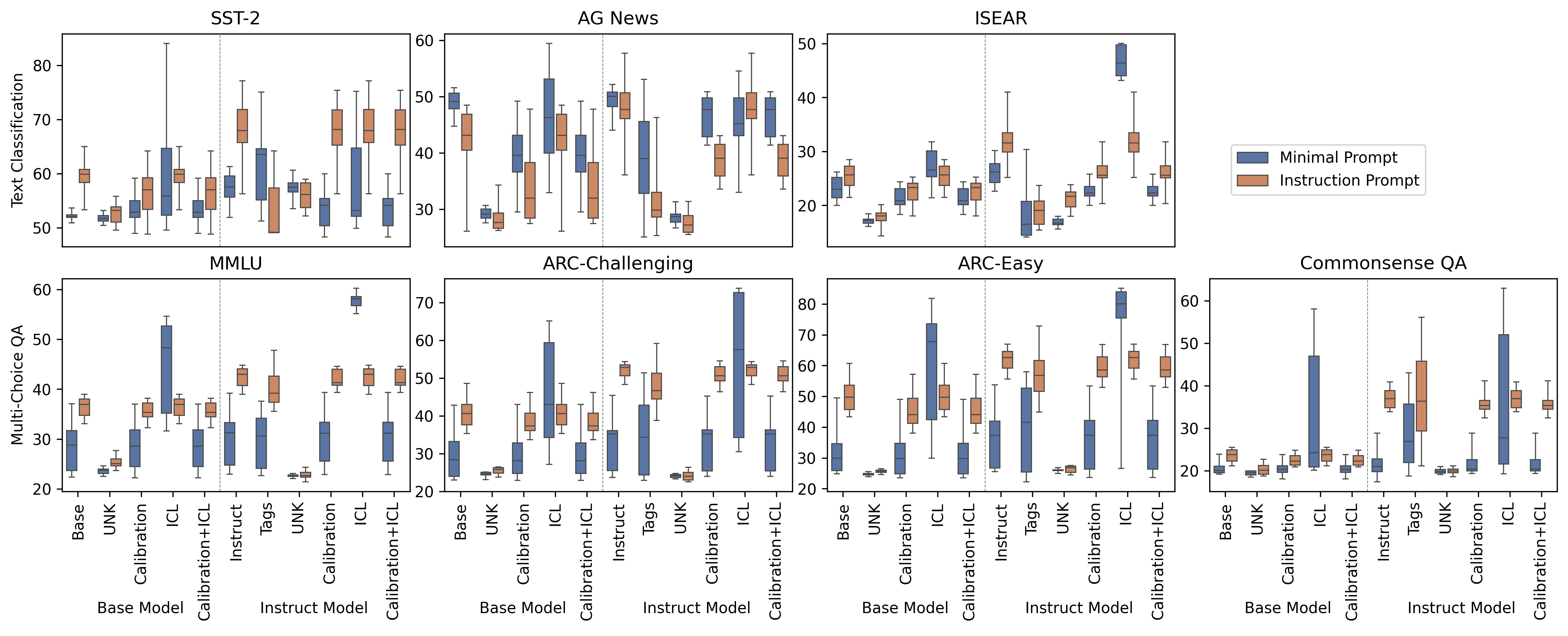}
        \subcaption{Gemma-3-4B}
    \end{subfigure}
    \hfill
    
    \caption{Sensitivity of Qwen-3-4B (a), and Gemma-3-4B (b) models to prompt variations measured as token probability accuracy and its deviation across different prompts. The problem of underspecification can be solved using a combination of instruction prompts and in-context learning.}
    \label{fig:main-results-further-models-qwen}
\end{figure*}

In this appendix, we provide results for the LLaMA-3.2-3B and LLaMA-2-13B models in Figure~\ref{fig:main-results-further-models} and for Qwen-3-4B and Gemma-3-4B in Figure~\ref{fig:main-results-further-models-qwen}. Compared to the main content of the paper, the observed results are not as straightforward, as discussed in Section~\ref{sec:emprirical-eval}. For the text classification datasets (SST2, AG News and ISEAR), the performance of the instruction prompts is sometimes lower and with higher deviation in results as compared to minimal prompts. Similar results can be observed for the MCQA datasets as well, although less frequently. These observations might be a result of lower capabilties of the smaller (LLaMA-3.2-3B) or older (LLaMA-2-13B) models that cannot as easily deal with the longer instructions. Once the instruction-tuned variants of the models are used, in-context learning or calibration, the instruction prompt formats are almost always better (higher performance and lower deviation) than minimal prompts. However, we still observe that the minimal prompt formats perform much better on the AG News dataset than the instruction formats, which might point to the specific of this dataset.

\subsection{Main Results: Generation Evaluation}
\label{app:main-results-generation}

\begin{figure*}[!th]
     \centering
     \begin{subfigure}[t]{.9\linewidth}
        \centering
        \includegraphics[width=1\linewidth]{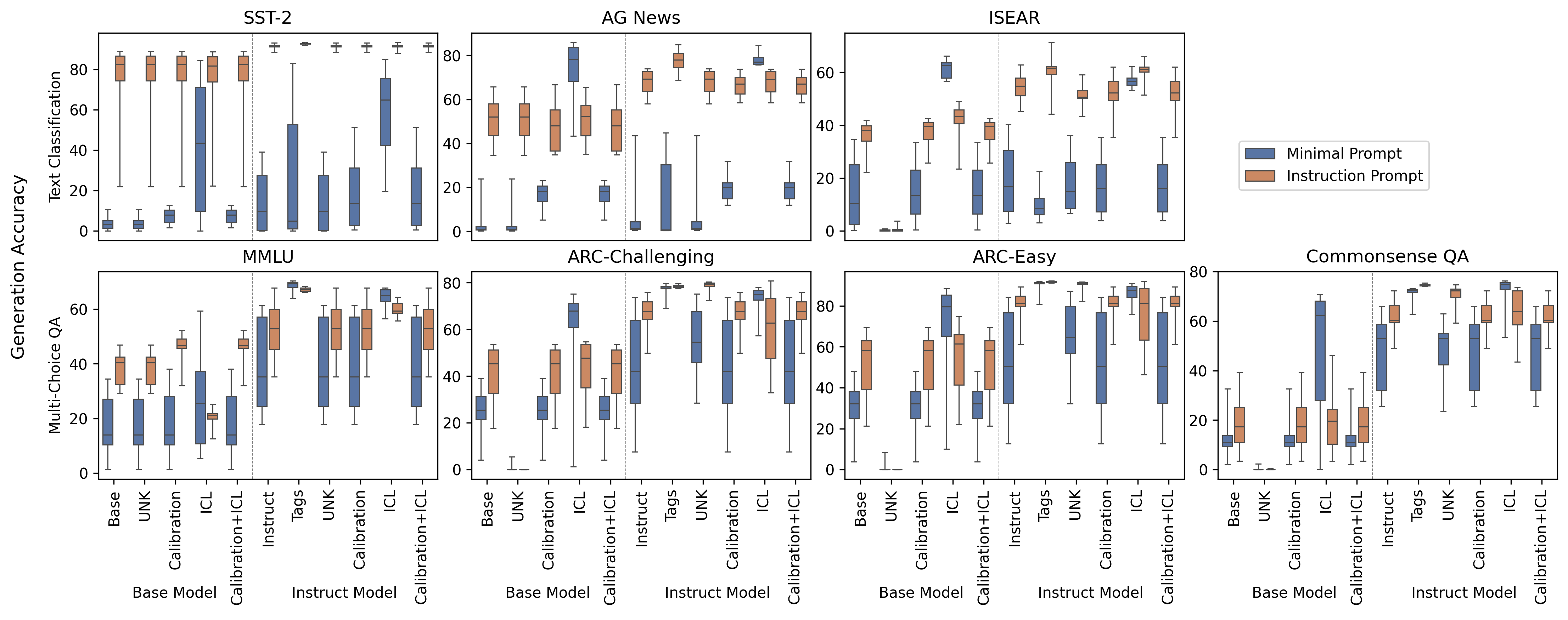}
        \subcaption{LLaMA-3.1-8B}
    \end{subfigure}
    \hfill
     \begin{subfigure}[t]{.9\linewidth}
        \centering
        \includegraphics[width=1\linewidth]{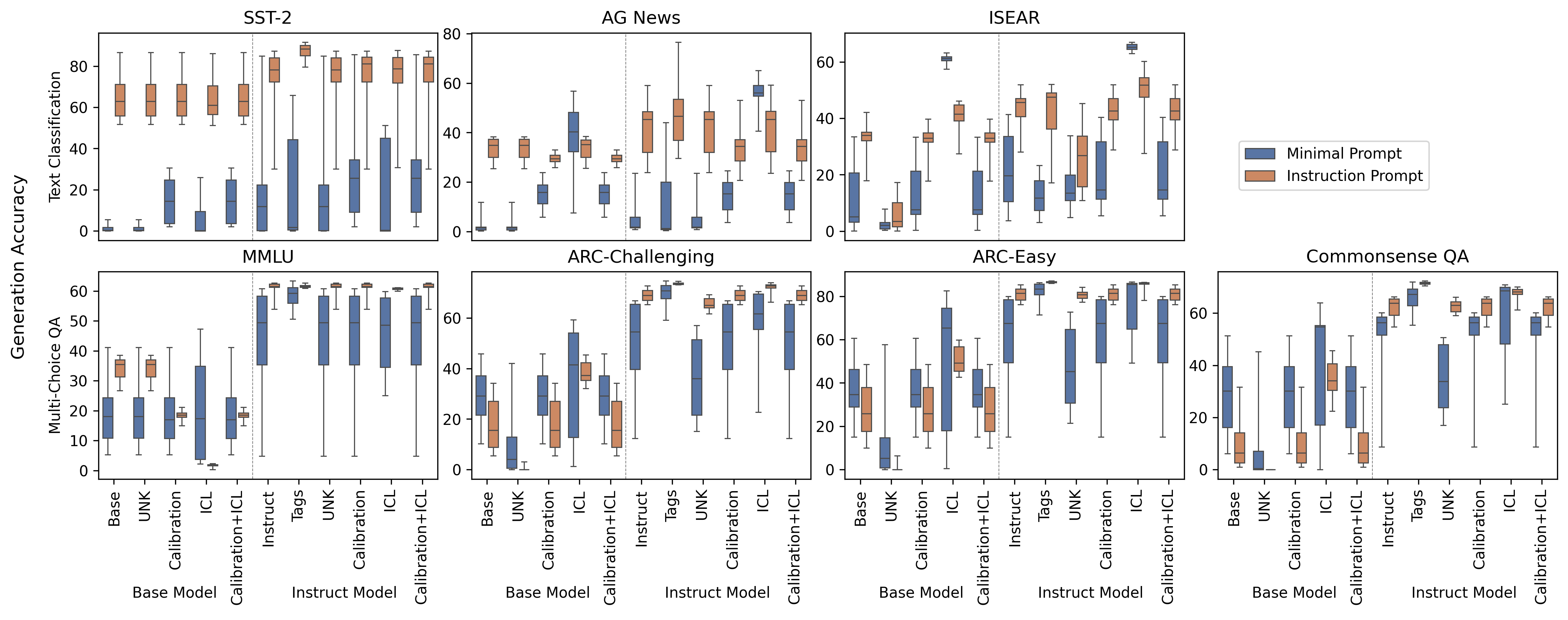}
        \subcaption{LLaMA-3.2-3B}
    \end{subfigure}
    \hfill
    \begin{subfigure}[t]{.9\linewidth}
        \centering
        \includegraphics[width=1\linewidth]{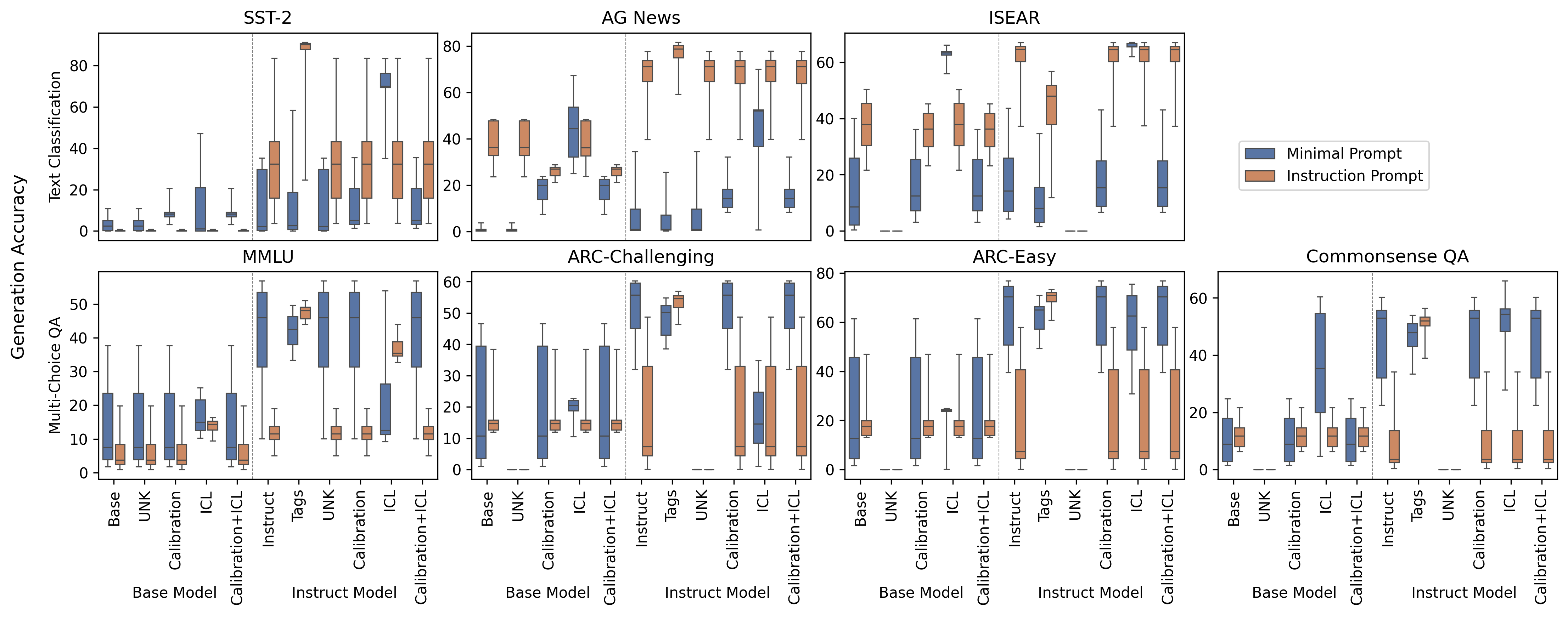}
        \subcaption{LLaMA-2-13B}
    \end{subfigure}
    \hfill
    
    \caption{Sensitivity of LLaMA-3.1-8B (a), LLaMA-3.2-3B (b), LLaMA-2-13B (c) models to prompt variations measured as generation accuracy and its deviation across different prompts. The problem of underspecification can be solved using a combination of instruction prompts and in-context learning.}
    \label{fig:main-results-generation}
\end{figure*}

\begin{figure*}[!th]
     \centering
    \begin{subfigure}[t]{.9\linewidth}
        \centering
        \includegraphics[width=1\linewidth]{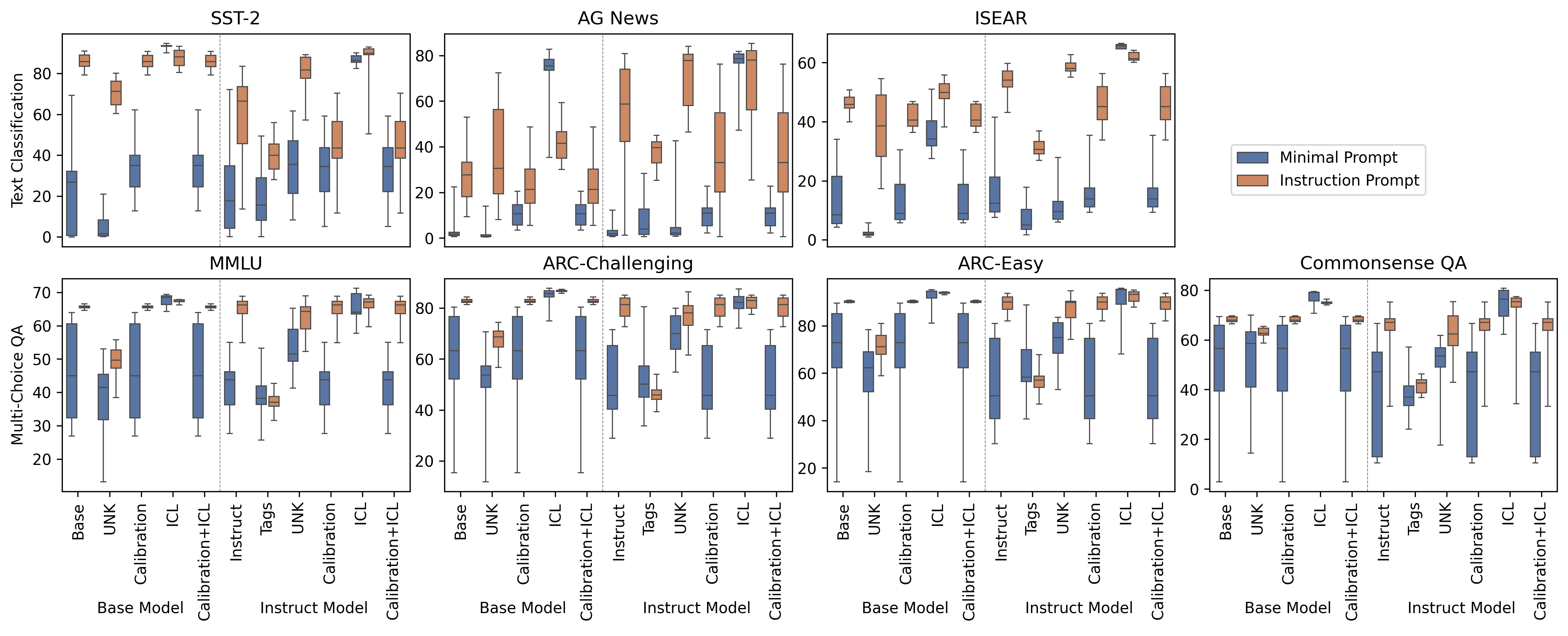}
        \subcaption{Qwen-3-4B}
    \end{subfigure}
    \hfill
    \begin{subfigure}[t]{.9\linewidth}
        \centering
        \includegraphics[width=1\linewidth]{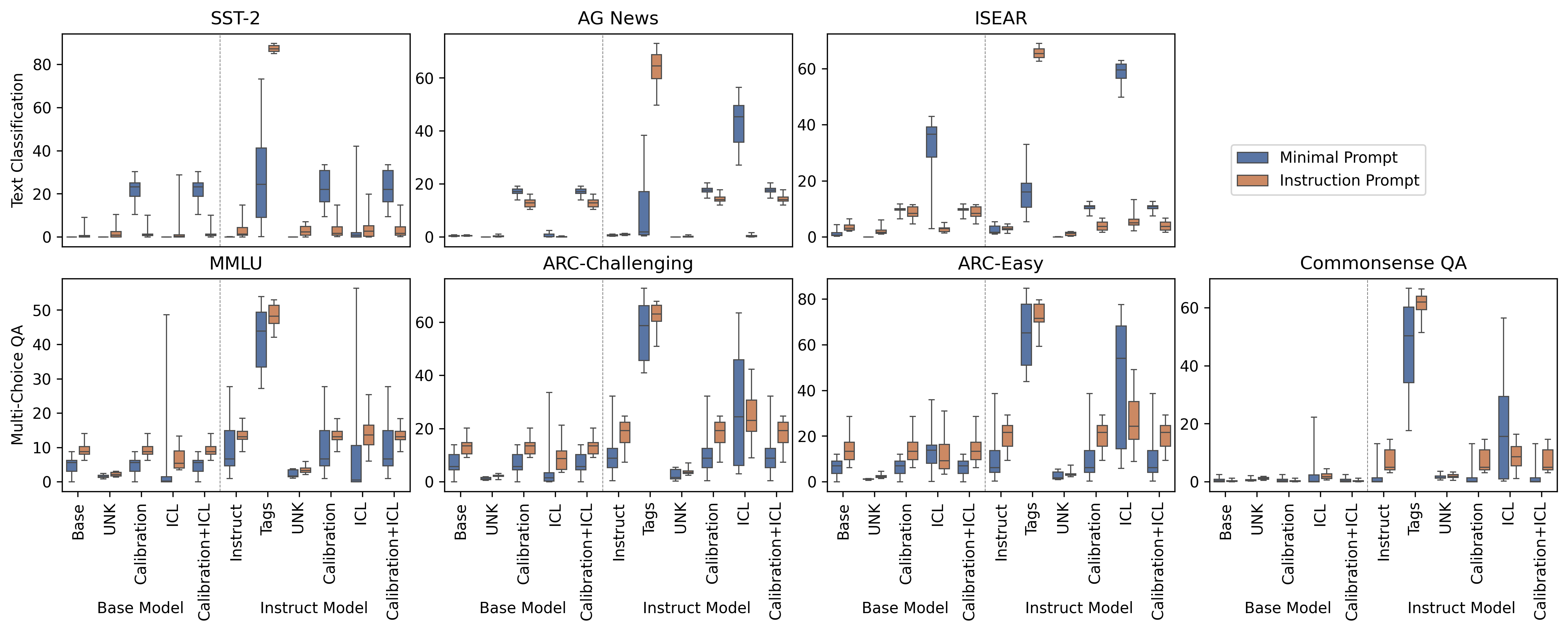}
        \subcaption{Gemma-3-4B}
    \end{subfigure}
    \hfill
    
    \caption{Sensitivity of Qwen-3-4B (a) and Gemma-3-4B (b) models to prompt variations measured as generation accuracy and its deviation across different prompts. The problem of underspecification can be solved using a combination of instruction prompts and in-context learning.}
    \label{fig:main-results-generation-qwen}
\end{figure*}

In this appendix, we provide visualisations for the generation evaluation of all models in Figure~\ref{fig:main-results-generation} and Figure~\ref{fig:main-results-generation-qwen}. Overall, we observe that without using any strategies for mitigating prompt sensitivity, the minimal prompts formats perform much worse, often times even being unable to outperform simple random chance. On the other hand, the performance on instruction prompt formats is consistent, although lower than when using label token probability evaluation. However, we still observe that even with instruction prompt formats, the LLaMA-3.2 (especially on MCQA) and LLaMA-2 models do not perform that well. These observations point to their lower capabilities in generation even though the probability distribution might be good enough to provide high performance.

\subsection{Label Token Probability: Additional Models}
\label{app:probabilities-other-models}

\begin{figure*}[!th]
     \centering
     \begin{subfigure}[t]{.9\linewidth}
        \centering
        \includegraphics[width=1\linewidth]{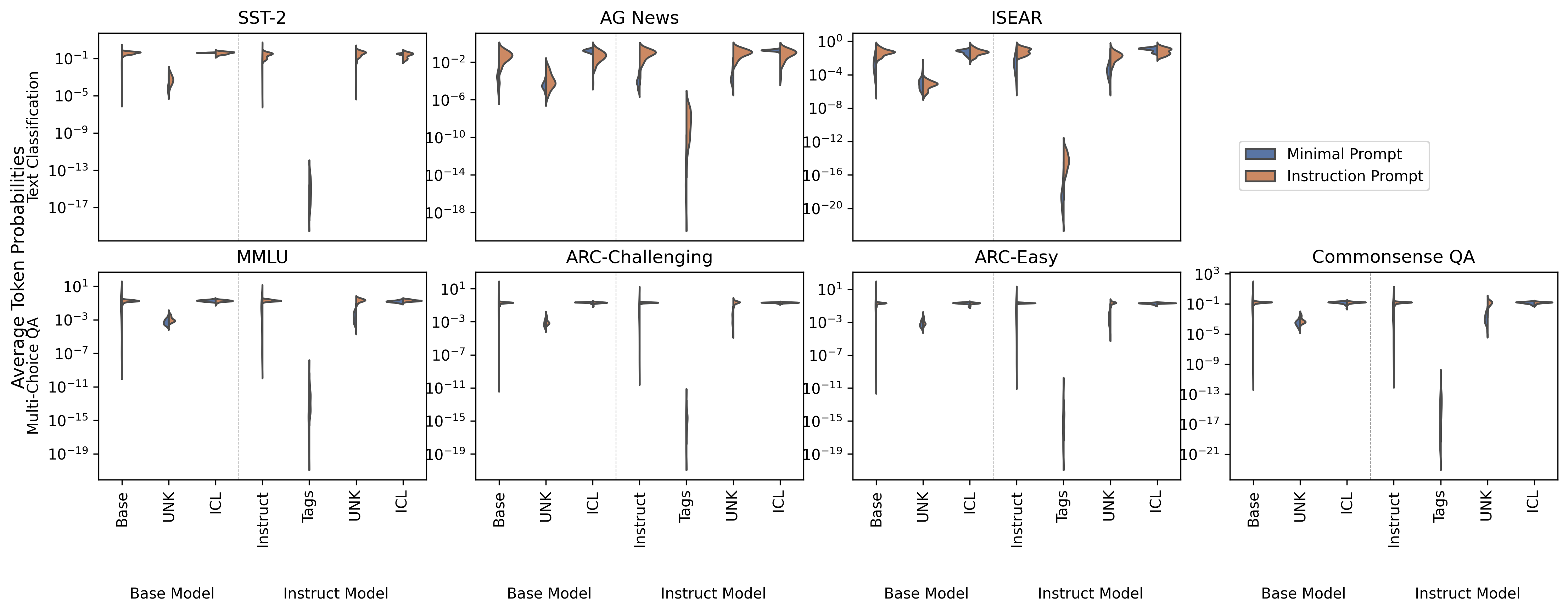}
        \subcaption{Qwen-3-4B}
    \end{subfigure}
    \hfill
    \begin{subfigure}[t]{.9\linewidth}
        \centering
        \includegraphics[width=1\linewidth]{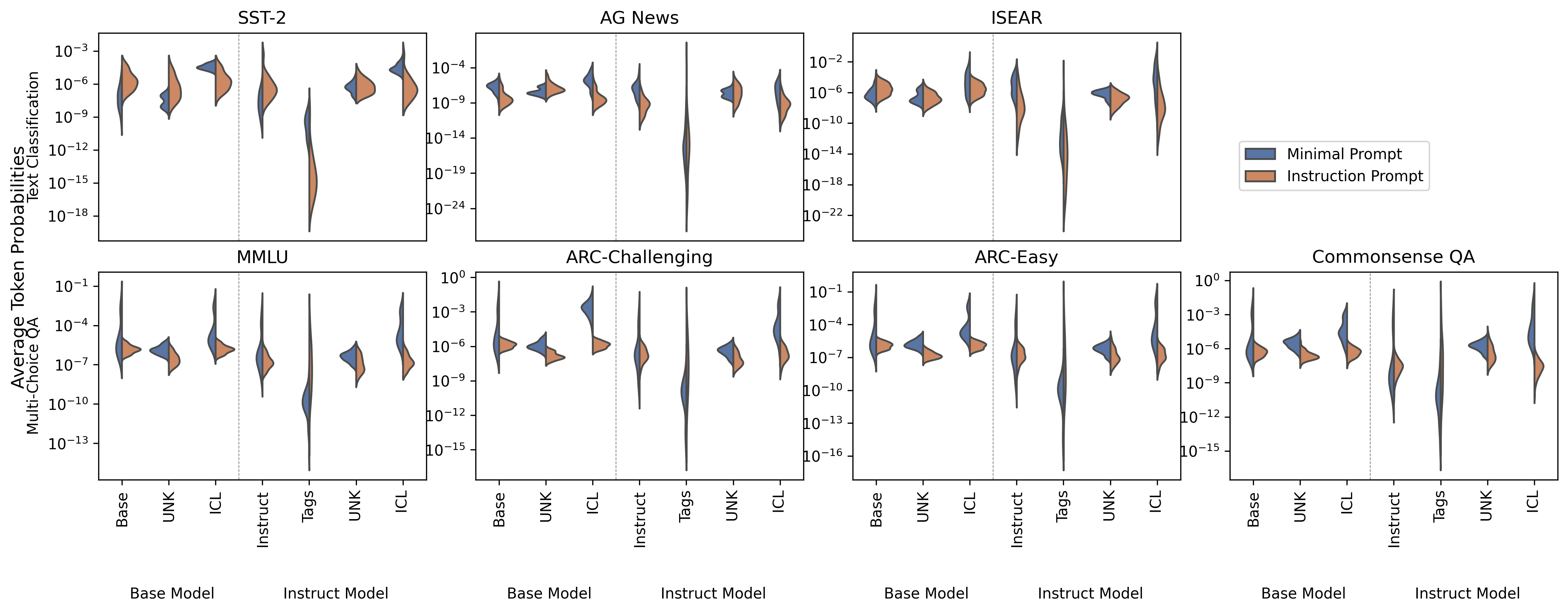}
        \subcaption{Gemma-3-4B}
    \end{subfigure}
    \hfill
    
    \caption{Token probability distribution across different prompts and individual class labels for Qwen-3-4B (a) and Gemma-3-4B (b) models. The instruction prompt formats and in-context learning consistently lead to highest label token probability values. On the other hand, the probabilities for minimal prompt formats are significantly smaller with larger distribution.}
    \label{fig:probability-additional-results-qwen}
\end{figure*}

In this appendix, we provide results for the label token probability analysis for the LLaMA-3.2-3B and LLaMA-2-13B models in Figure~\ref{fig:probe-additional-results} and for Qwen-3-4B and Gemma-3-4B in Figure~\ref{fig:probability-additional-results-qwen}. Overall, the results follow the findings described in the main content of the paper (Section~\ref{sec:logit-analysis}). The only difference we can observe is degradation of the label token probability distributions on some datasets (AG News) for the LLaMA-3.2-3B model.

\begin{figure*}[!th]
     \centering
     \begin{subfigure}[t]{.9\linewidth}
        \centering
        \includegraphics[width=1\linewidth]{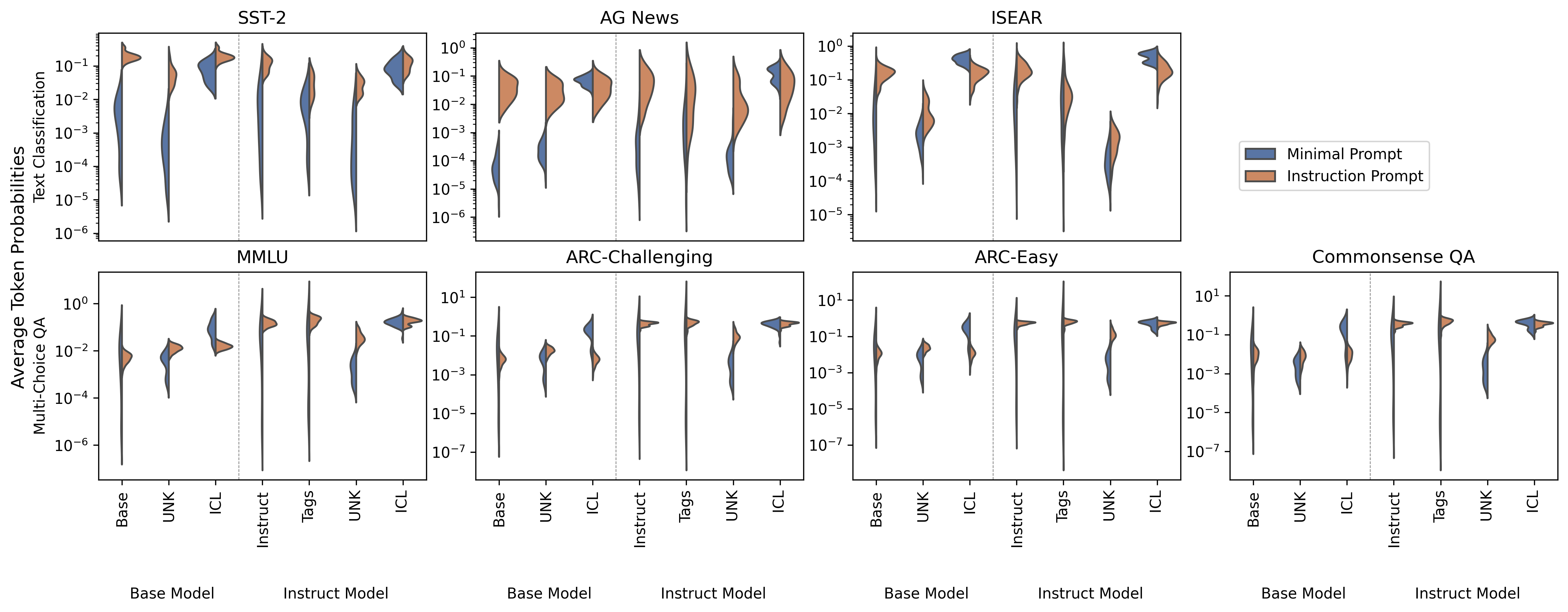}
        \subcaption{LLaMA-3.2-3B}
    \end{subfigure}
    \hfill
    \begin{subfigure}[t]{.9\linewidth}
        \centering
        \includegraphics[width=1\linewidth]{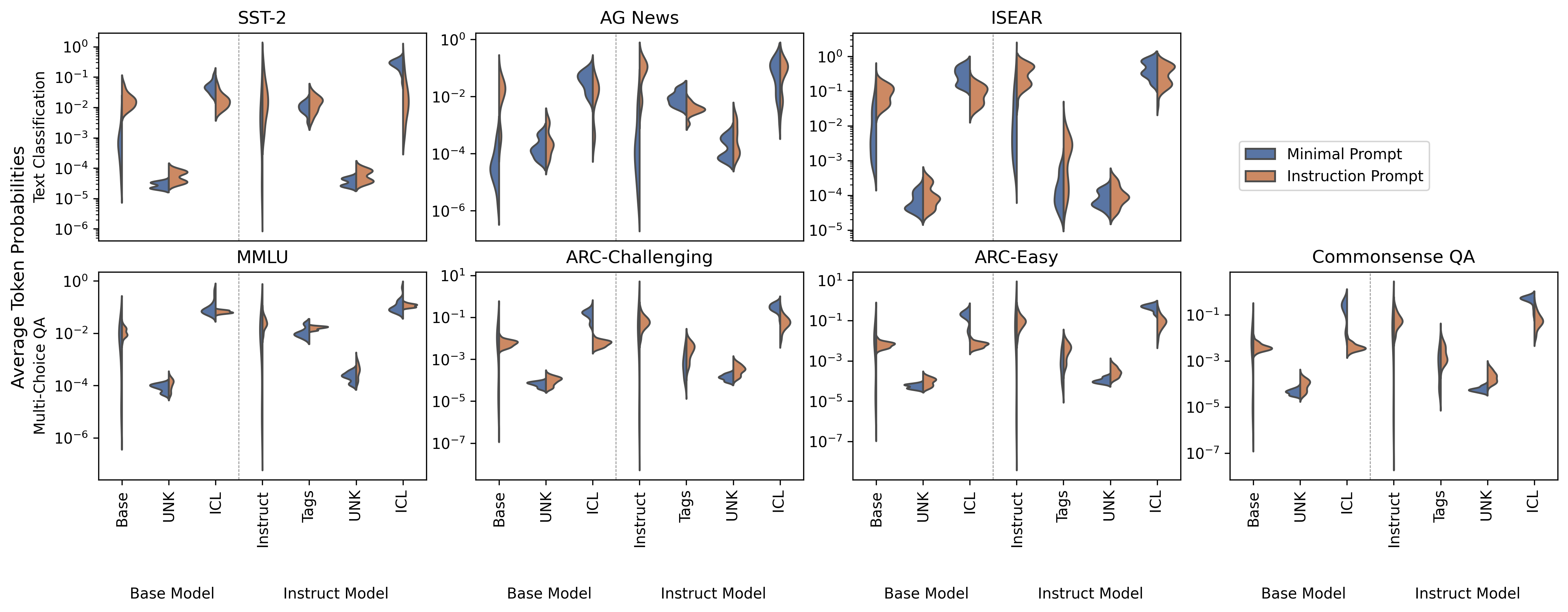}
        \subcaption{LLaMA-2-13B}
    \end{subfigure}
    \hfill
    
    \caption{Token probability distribution across different prompts and individual class labels for LLaMA-3.2-3B (a) and LLaMA-2-13B (b) models. The instruction prompt formats and in-context learning consistently lead to highest label token probability values. On the other hand, the probabilities for minimal prompt formats are significantly smaller with larger distribution.}
    \label{fig:probability-additional-results}
\end{figure*}

\subsection{Linear Probe Analysis}
\label{app:linear-probe-analysis}

In this appendix we provide further information and results for the linear probe analysis from the main content of the paper. To obtain the results, we train a separate SVM classifier with RBF kernel on each layer of the model. To train the classifier we follow these steps:
\begin{enumerate}
    \item For the given dataset, we take the training set with all the samples.
    \item For the given prompt, we take all of the training samples and use them to construct the prompt representation of the sample.
    \item All of these samples are passed to the LLM and the outputs of each layer are collected.
    \item After processing all of the samples, we train a separate SVM model on each layer, using the ouputs (e.g., activations) as features and the dataset labels as targets.
    \item After training the model, we repeat the same for the test set, using the SVM only to classify each sample.
    \item The accuracy of the SVM on each layer is calculated as the mean accuracy across these samples.
    \item The whole process is repeated for each prompt and dataset and then used to visualise the results.
\end{enumerate}

\begin{figure*}[!th]
     \centering
     \begin{subfigure}[t]{.9\linewidth}
        \centering
        \includegraphics[width=1\linewidth]{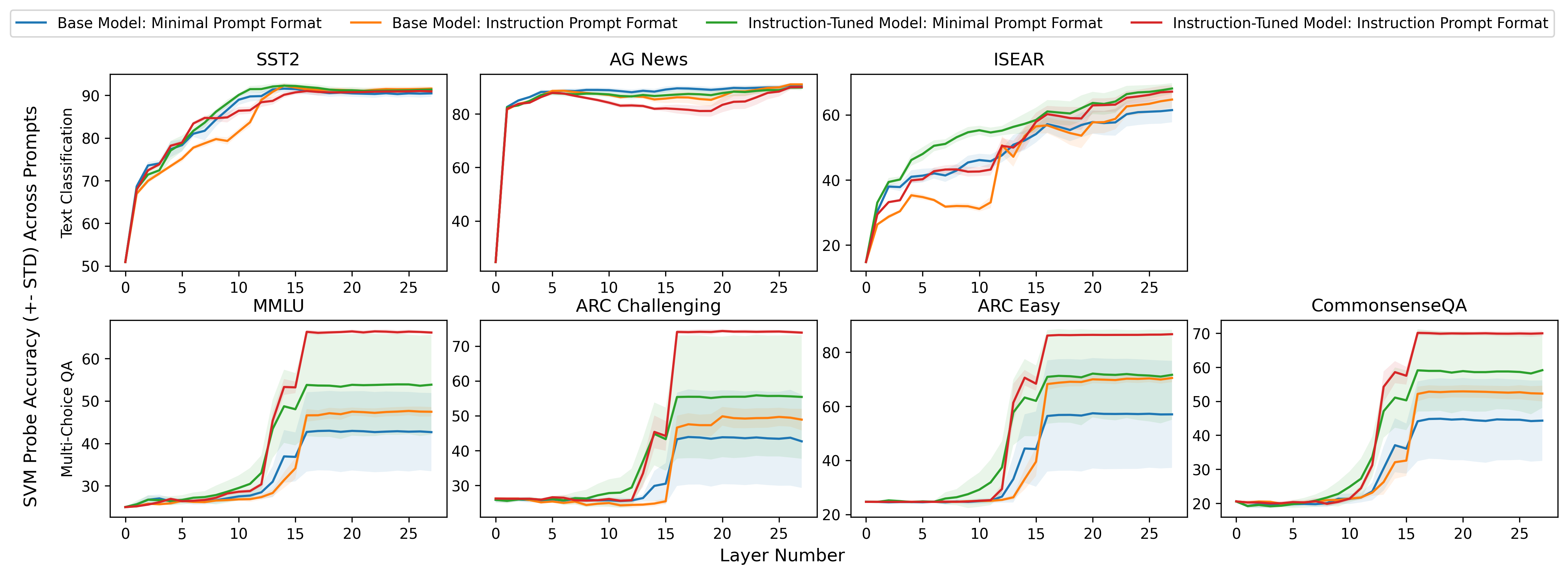}
        \subcaption{LLaMA-3.2-3B}
    \end{subfigure}
    \hfill
    \begin{subfigure}[t]{.9\linewidth}
        \centering
        \includegraphics[width=1\linewidth]{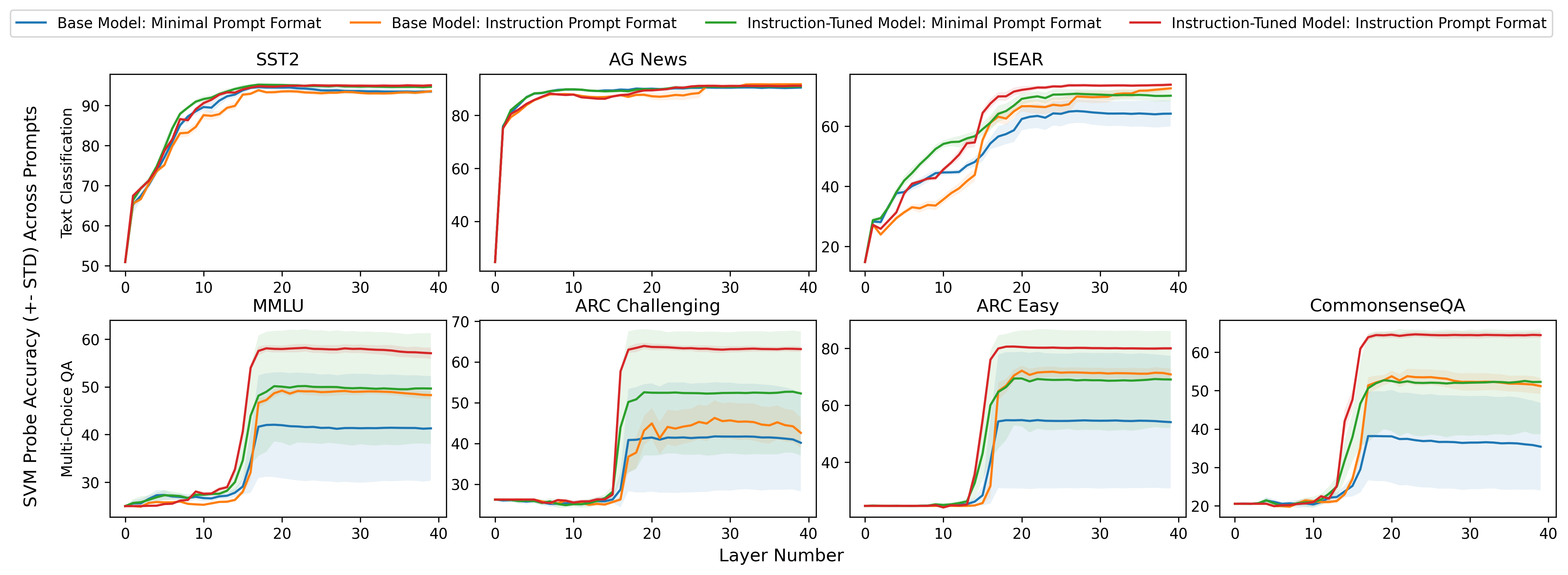}
        \subcaption{LLaMA-2-13B}
    \end{subfigure}
    \hfill
    
    \caption{The mean accuracy of the SVM model over different minimal and instruction prompt formats for all layers of the LLaMA-3.2-3B (a) and LLaMA-2-13B (b) models. Minimal prompts affect the internal representations and the model label decodability, especially for multi-choice QA datasets.}
    \label{fig:probe-additional-results}
\end{figure*}

\begin{figure*}[!th]
     \centering
     \begin{subfigure}[t]{.9\linewidth}
        \centering
        \includegraphics[width=1\linewidth]{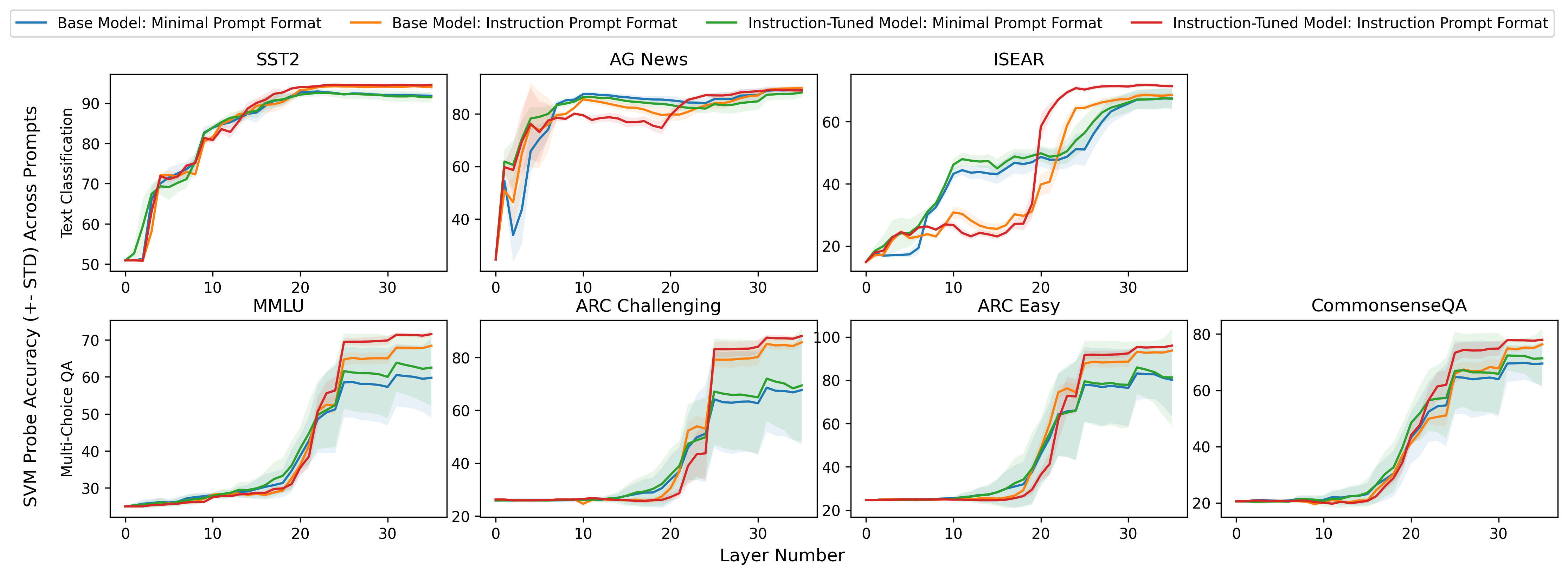}
        \subcaption{Qwen-3-4B}
    \end{subfigure}
    \hfill
    \begin{subfigure}[t]{.9\linewidth}
        \centering
        \includegraphics[width=1\linewidth]{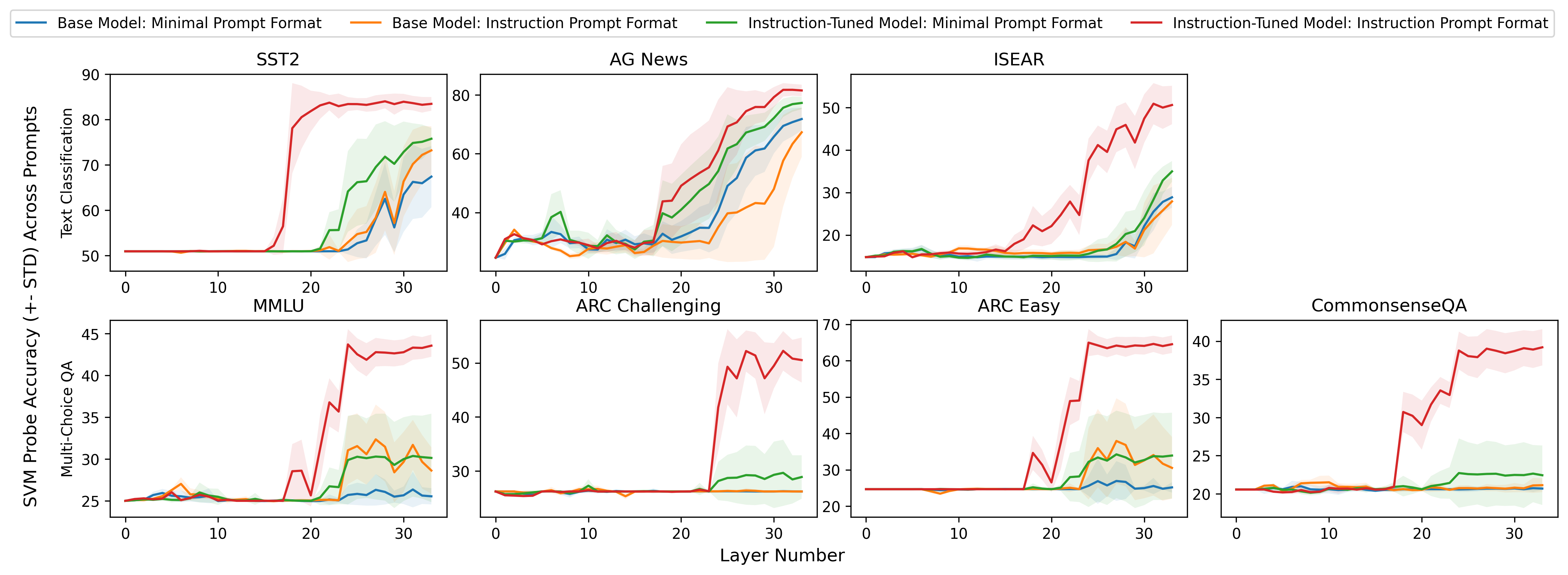}
        \subcaption{Gemma-3-4B}
    \end{subfigure}
    \hfill
    
    \caption{The mean accuracy of the SVM model over different minimal and instruction prompt formats for all layers of the Qwen-3-4B (a) and Gemma-3-4B (b) models. Minimal prompts affect the internal representations and the model label decodability, especially for multi-choice QA datasets.}
    \label{fig:probe-additional-results-qwen}
\end{figure*}

The results for the LLaMA-3.2-3B and LLaMA-2-13B models are included in Figure~\ref{fig:probe-additional-results} and for Qwen-3-4B and Gemma-3-4B in Figure~\ref{fig:probe-additional-results-qwen}. Compared to the main body of the paper, we do not observe any significant differences.

\begin{figure*}[!th]
     \centering
     \begin{subfigure}[t]{.9\linewidth}
        \centering
        \includegraphics[width=1\linewidth]{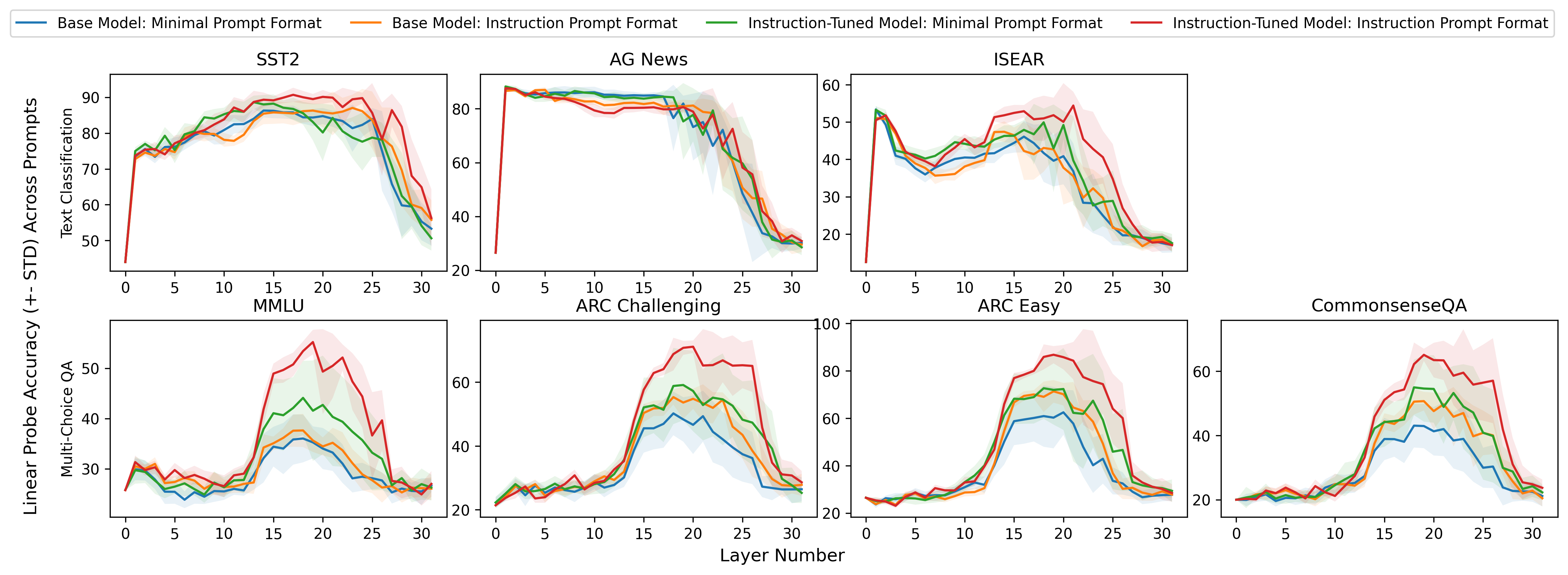}
        \subcaption{LLaMA-3.1-8B}
    \end{subfigure}
    \hfill
     \begin{subfigure}[t]{.9\linewidth}
        \centering
        \includegraphics[width=1\linewidth]{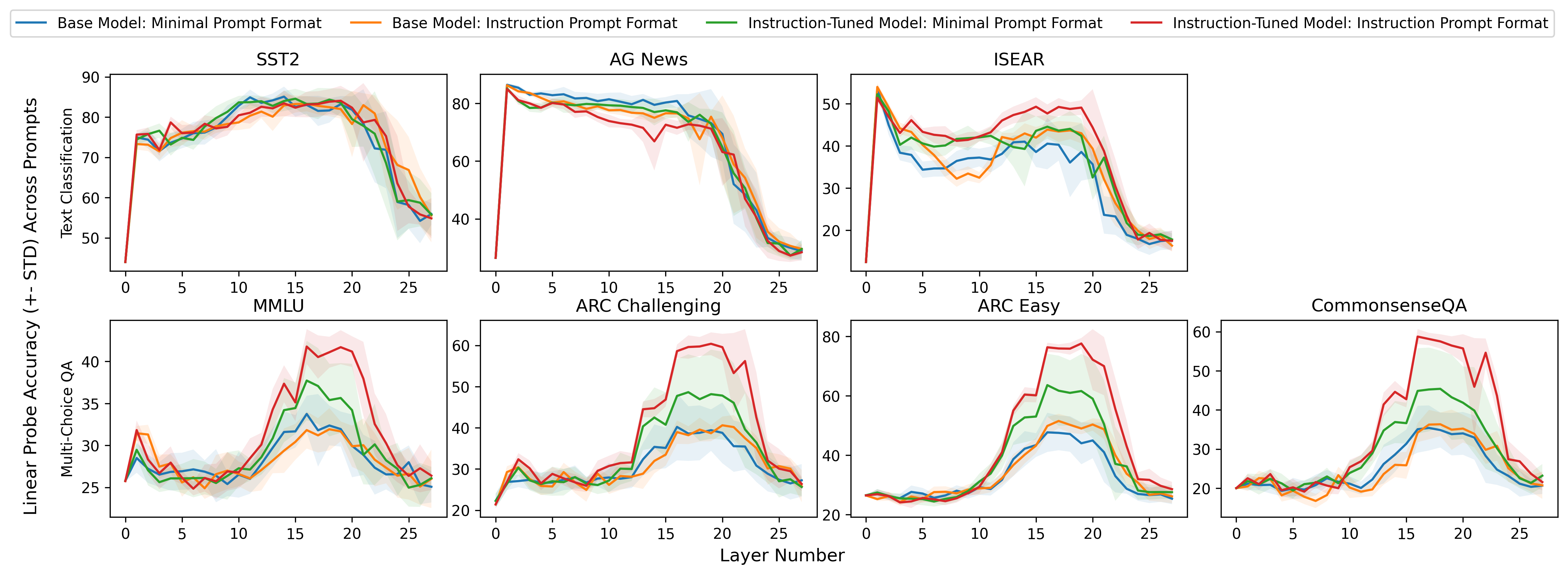}
        \subcaption{LLaMA-3.2-3B}
    \end{subfigure}
    \hfill
    \begin{subfigure}[t]{.9\linewidth}
        \centering
        \includegraphics[width=1\linewidth]{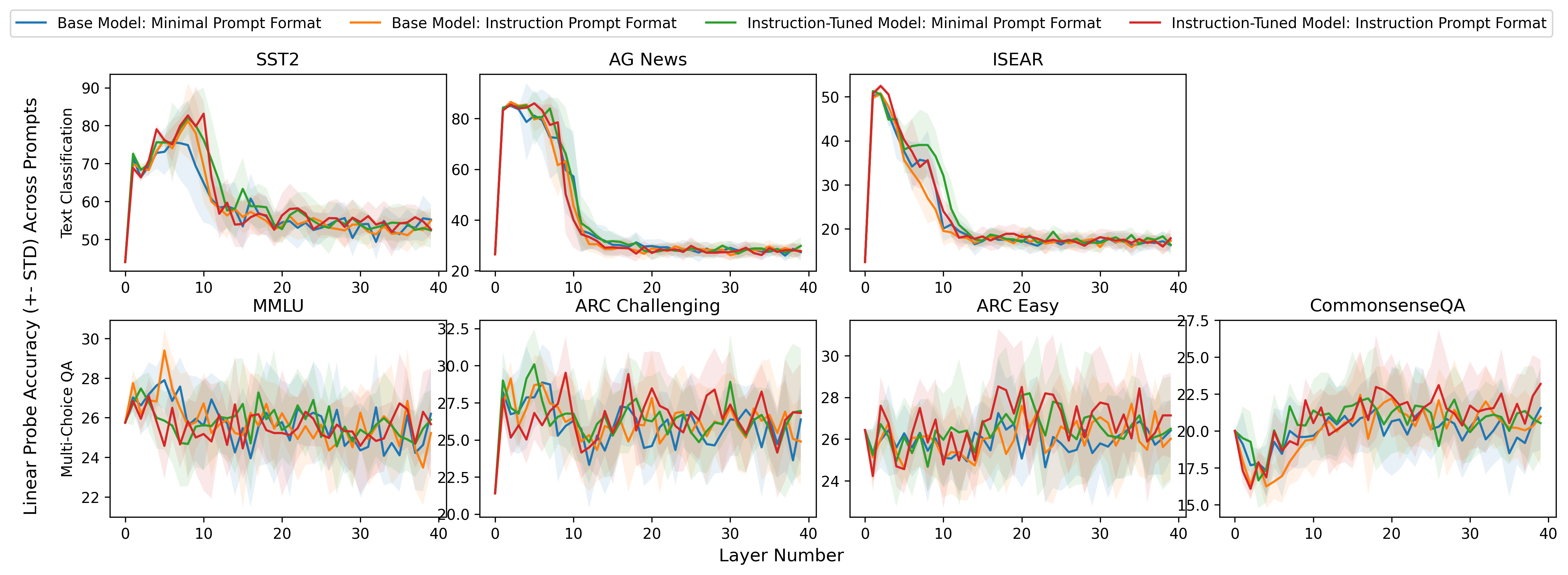}
        \subcaption{LLaMA-2-13B}
    \end{subfigure}
    \hfill
    
    \caption{The mean accuracy of the linear regression model with Ridge regression over different minimal and instruction prompt formats for all layers of all models. We can observe that the linear model cannot handle the possibly non-linear representations in the later layers, leading to degraded performance. However, we can still observe that minimal prompts affect the internal representations and the model label decodability, especially for multi-choice QA datasets.}
    \label{fig:linear-probe-results}
\end{figure*}

\textbf{Why we use SVM model with RBF kernel instead of simple linear regression model?} In our preliminary experiments, we have trained linear regression with Ridge regression instead of the SVM model. However, in this case, we have observed a curious behaviour of the model (as illustrated in Figure~\ref{fig:linear-probe-results}). For the first half of the network, we have observed the same behaviour as with the SVM model. However, towards the later layers, the performance of the linear model have significantly degraded, performing worse than random chance. We have hypothesised that the reason for this is that the representations in the later layers are of non-linear characters (or more specifically the mapping between the activations and the final labels in the label decodability task is non-linear) which the linear models cannot deal with. For this reason, we have opted for the SVM model with RBF kernel (that can handle non-linear representations) that confirmed our hypothesis. However, we include the results from the linear model as well for completeness sake.

\section{Experimental Setup Details}
\label{app:experimental-details}

The analysis and experiments are done on 7 English-only datasets of various characteristics and complexity, including the binary sentiment classification SST2~\citep{socher-etal-2013-recursive} dataset, the topic classification AG News~\citep{zhang2015agnews} dataset, emotion detection dataset ISEAR~\citep{scherer1994evidence-isear} and the multi-choice question answering datasets MMLU~\citep{hendrycks2021measuring}, ARC~\citep{allenai:arc} and CommonsenseQA~\citep{talmor-etal-2019-commonsenseqa}. The prompt formats used for each dataset are in Table~\ref{tab:prompt-format-sst2} for SST2, Table~\ref{tab:prompt-format-agnews} for the AG News, Table~\ref{tab:prompt-format-isear} for the ISEAR dataset, Table~\ref{tab:prompt-format-mmlu} for MMLU and ARC, and Table~\ref{tab:prompt-format-cqa} for the CommonsenseQA dataset. For generating new prompt format (when necessary) we use prompt format specified in Table~\ref{tab:generation-prompt}. When performing any choices or evaluations in the pilot experiments (e.g., finding the best performing calibration strategy), we always use validation sets of the datasets, or in case they are not available, we use 20\% of the train dataset for the decisions (in those cases, only the remaining 80\% of the dataset are used for any analysis that required training sets).

\begin{table*}[!tbh]
\centering
\small
\caption{Prompt formats for the SST2 dataset.}
\label{tab:prompt-format-sst2}
\begin{tabularx}{\textwidth}{@{}m{0.02cm}X@{}}
\toprule
 & \textbf{Prompt Text} \\ \midrule
\multirow{10}{*}{\rotatebox{90}{\begin{footnotesize}\textbf{Minimal Prompt Format}\end{footnotesize}}}   & "\{sample\}\textbackslash nMy feedback to the film is "\\
& "\{sample\}\textbackslash nOverall, my feedback to the film is " \\
& "\{sample\}\textbackslash nConsidering the details provided, my emotional reaction is " \\
& "\{sample\}\textbackslash nReflecting on the content, my emotional stance is " \\
& "\{sample\}\textbackslash nGiven the information above, my sentiment evaluation is " \\
& "\{sample\}\textbackslash nAnalyzing the feedback, my emotional assessment is " \\
& "\{sample\}\textbackslash nBased on the review, my overall sentiment impression is " \\
& "\{sample\}\textbackslash nAfter thoroughly considering the review, my sentiment perspective is " \\
& "\{sample\}\textbackslash nAll in all, the film was " \\
& "\{sample\}\textbackslash nIn summary, the film was " \\ \midrule  

\multirow{19}{*}{\rotatebox{90}{\begin{footnotesize}\textbf{Instruction Prompt Format}\end{footnotesize}}}

& "Determine sentiment of the sentence using following options: negative; positive. Use only these two options.\textbackslash nSentence: \{sample\}\textbackslash nSentiment: " \\
& "Your task is to determine the sentiment of a given sentence. Use only following options: negative; positive.\textbackslash nSentence: \{sample\}\textbackslash nSentiment: " \\
& "Classify the sentiment expressed in the sentence. Choose one of the following: negative; positive.\textbackslash nSentence: \{sample\}\textbackslash nSentiment: " \\
& "Analyze the sentiment of the provided sentence and select the appropriate label: negative; positive. Use only these two options.\textbackslash nSentence: \{sample\}\textbackslash nSentiment: " \\
& "Evaluate the tone of the sentence and determine if it is: negative or positive. Use only these two options.\textbackslash nSentence: \{sample\}\textbackslash nSentiment: " \\
& "Your goal is to assess the sentiment of the sentence and respond using one of these categories: negative; positive. Use only these two options.\textbackslash nSentence: \{sample\}\textbackslash nSentiment: " \\
& "Given a sentence, identify its sentiment and assign one of the following labels: negative; positive. Use only these two options.\textbackslash nSentence: \{sample\}\textbackslash nSentiment: " \\
& "Decide whether the sentiment of the sentence is: negative or positive. Choose only one.\textbackslash nSentence: \{sample\}\textbackslash nSentiment: " \\
& "Identify the overall sentiment conveyed in the sentence. Use only these labels: negative; positive.\textbackslash nSentence: \{sample\}\textbackslash nSentiment: " \\
& "Determine if the sentiment in the sentence is best described as: negative or positive. Use only these two options.\textbackslash nSentence: \{sample\}\textbackslash nSentiment: " \\

\bottomrule

\end{tabularx}
\end{table*}

\begin{table*}[!tbh]
\centering
\small
\caption{Prompt formats for the AG News dataset.}
\label{tab:prompt-format-agnews}
\begin{tabularx}{\textwidth}{@{}m{0.02cm}X@{}}
\toprule
 & \textbf{Prompt Text} \\ \midrule
\multirow{10}{*}{\rotatebox{90}{\begin{footnotesize}\textbf{Minimal Prompt Format}\end{footnotesize}}}   & "\{sample\}\textbackslash nThis topic is about "\\
& "\{sample\}\textbackslash nThe label that best describes this news article is " \\
& "\{sample\}\textbackslash nThis piece of news is regarding " \\
& "\{sample\}\textbackslash nThe news article is about " \\
& "\{sample\}\textbackslash nCentral themes of this news piece encompass " \\
& "\{sample\}\textbackslash nThe central theme of this article revolves around " \\
& "\{sample\}\textbackslash nIt can be labeled as " \\
& "\{sample\}\textbackslash nIts category is " \\
& "\{sample\}\textbackslash nIn this article, it talks about " \\
& "\{sample\}\textbackslash nI think the news can be classified as " \\ \midrule  

\multirow{19}{*}{\rotatebox{90}{\begin{footnotesize}\textbf{Instruction Prompt Format}\end{footnotesize}}}

& "Determine topic of the sentence using following options: world; sports; business; technology.\textbackslash nSentence: \{sample\}\textbackslash nTopic: " \\
& "Your task is to determine the topic of a given sentence. Use only following options: world; sports; business; technology.\textbackslash nSentence: \{sample\}\textbackslash nTopic: " \\
& "Classify the topic expressed in the sentence. Choose one of the following: world; sports; business; technology.\textbackslash nSentence: \{sample\}\textbackslash nTopic: " \\
& "Analyze the topic of the provided sentence and select the appropriate label: world; sports; business; technology.\textbackslash nSentence: \{sample\}\textbackslash nTopic: " \\
& "Evaluate the topic of the sentence and determine if it is: world; sports; business; technology.\textbackslash nSentence: \{sample\}\textbackslash nTopic: " \\
& "Your goal is to assess the topic of the sentence and respond using one of these categories: world; sports; business; technology.\textbackslash nTopic: \{sample\}\textbackslash nTopic: " \\
& "Decide whether the topic of the sentence is: world; sports; business; technology. Choose only one.\textbackslash nSentence: \{sample\}\textbackslash nTopic: " \\
& "Label the topic of this sentence. The only acceptable options are: world; sports; business; technology.\textbackslash nSentence: \{sample\}\textbackslash nTopic: " \\
& "Using only predefined labels, identify whether the topic is: world; sports; business; technology.\textbackslash nSentence: \{sample\}\textbackslash nTopic: " \\
& "As part of this annotation task, determine whether the topic sentence is: world; sports; business; technology.\textbackslash nSentence: \{sample\}\textbackslash nTopic: " \\

\bottomrule

\end{tabularx}
\end{table*}

\begin{table*}[!tbh]
\centering
\small
\caption{Prompt formats for the ISEAR dataset.}
\label{tab:prompt-format-isear}
\begin{tabularx}{\textwidth}{@{}m{0.02cm}X@{}}
\toprule
 & \textbf{Prompt Text} \\ \midrule
\multirow{10}{*}{\rotatebox{90}{\begin{footnotesize}\textbf{Minimal Prompt Format}\end{footnotesize}}}   & "\{sample\}\textbackslash nIt seems to evoke a feeling of " \\
& "\{sample\}\textbackslash nIt can be interpreted as expressing " \\
& "\{sample\}\textbackslash nThe underlying emotion seems to be " \\
& "\{sample\}\textbackslash nFeeling-wise, this comes across as " \\
& "\{sample\}\textbackslash nThe emotional tone here is " \\
& "\{sample\}\textbackslash nThis story is imbued with " \\
& "\{sample\}\textbackslash nReading this, one might feel " \\
& "\{sample\}\textbackslash nThis piece stirs up " \\
& "\{sample\}\textbackslash nEmotionally, this is " \\ 
& "\{sample\}\textbackslash nThis depicts "\\ \midrule  

\multirow{19}{*}{\rotatebox{90}{\begin{footnotesize}\textbf{Instruction Prompt Format}\end{footnotesize}}}

& "Determine the emotion of the sentence using following options: shame; joy; sadness; disgust; anger; fear; guilt. Use only these options.\textbackslash nSentence: \{sample\}\textbackslash nEmotion: " \\
& "Your task is to determine the emotion expressed in a given sentence. Use only following options: shame; joy; sadness; disgust; anger; fear; guilt.\textbackslash nSentence: \{sample\}\textbackslash nEmotion: " \\
& "Classify the emotion expressed in the sentence. Choose one of the following: shame; joy; sadness; disgust; anger; fear; guilt.\textbackslash nSentence: \{sample\}\textbackslash nEmotion: " \\
& "Analyse the emotion of the provided sentence and select the appropriate label: shame; joy; sadness; disgust; anger; fear; guilt. Use only these options.\textbackslash nSentence: \{sample\}\textbackslash nEmotion: " \\
& "Evaluate the emotional tone of the sentence and determine which category applies: shame; joy; sadness; disgust; anger; fear; guilt. Use only these options.\textbackslash nSentence: \{sample\}\textbackslash nEmotion: " \\
& "Your goal is to assess the emotion of the sentence and respond using one of these categories: shame; joy; sadness; disgust; anger; fear; guilt. Use only these options.\textbackslash nTopic: \{sample\}\textbackslash nEmotion: " \\
& "Given a sentence, identify its emotion and assign one of the following labels: shame; joy; sadness; disgust; anger; fear; guilt. Use only these options.\textbackslash nSentence: \{sample\}\textbackslash nEmotion: " \\
& "Decide which emotion best describes the sentence: shame; joy; sadness; disgust; anger; fear; guilt. Choose only one.\textbackslash nSentence: \{sample\}\textbackslash nEmotion: " \\
& "Identify the primary emotion conveyed in the sentence. Use only these labels: shame; joy; sadness; disgust; anger; fear; guilt.\textbackslash nSentence: \{sample\}\textbackslash nEmotion: " \\
& "Determine if the emotion in the sentence is best described as one of the following: shame; joy; sadness; disgust; anger; fear; guilt. Use only these options.\textbackslash nSentence: \{sample\}\textbackslash nEmotion: " \\

\bottomrule

\end{tabularx}
\end{table*}

\begin{table*}[!tbh]
\centering
\small
\caption{Prompt formats for the MMLU and ARC datasets.}
\label{tab:prompt-format-mmlu}
\begin{tabularx}{\textwidth}{@{}m{0.02cm}X@{}}
\toprule
 & \textbf{Prompt Text} \\ \midrule
\multirow{10}{*}{\rotatebox{90}{\begin{footnotesize}\textbf{Minimal Prompt Format}\end{footnotesize}}}   & "Here is a question with several possible answers: \{sample\}\textbackslash n"\\
& "Review the choices below and determine the best fit: \{sample\}\textbackslash n" \\
& "Could you provide a response to the following question:\textbackslash n\{sample\}\textbackslash nAnswer: " \\
& "Please answer the following question:\textbackslash n\{sample\}\textbackslash nAnswer: " \\
& "\{sample\}\textbackslash nI think it is " \\
& "\{sample\}\textbackslash nIt is " \\
& "\{sample\}\textbackslash nConsidering everything, I think it is " \\
& "\{sample\}\textbackslash nThe best answer seems to be " \\
& "\{sample\}\textbackslash nEverything points to " \\
& "\{sample\}\textbackslash n" \\ \midrule  

\multirow{19}{*}{\rotatebox{90}{\begin{footnotesize}\textbf{Instruction Prompt Format}\end{footnotesize}}}

& "Select the best answer from the given options. Respond only with the letter corresponding to the correct choice.\textbackslash n\{sample\}\textbackslash nAnswer: " \\
& "Choose the most appropriate answer among the given options. Return only the letter of the selected choice, without any explanation or additional text.\textbackslash n\{sample\}\textbackslash nAnswer: " \\
& "From the listed answer choices, select the most accurate one. Reply only with the letter of your selection (A, B, C, or D). Do not include any extra commentary or reasoning.\textbackslash n\{sample\}\textbackslash nAnswer: " \\
& "Identify the correct answer from the provided options. Your response must consist of a single uppercase letter (A, B, C, or D) matching your chosen answer. No explanations are needed.\textbackslash n\{sample\}\textbackslash nAnswer: " \\
& "Pick the best option and reply with only the corresponding letter (A, B, C, or D).\textbackslash n\{sample\}\textbackslash nAnswer: " \\
& "After reading the question and the provided answer choices, choose the single best answer. Your reply must be only the letter (A, B, C, or D) associated with the correct choice.\textbackslash n\{sample\}\textbackslash nAnswer: " \\
& "Treat this like a multiple-choice exam. Choose the correct answer and reply with only the letter of the option you select (A, B, C, or D). No explanations are required.\textbackslash n\{sample\}\textbackslash nAnswer: " \\
& "Select the correct option. Your entire response must consist of a single capital letter corresponding to your choice (A, B, C, or D) — no extra words, punctuation, or reasoning.\textbackslash n\{sample\}\textbackslash nAnswer: " \\
& "Pick the answer you think is best. Just send back the letter (like "A", "B", "C" or "D") — no need to explain why.\textbackslash n\{sample\}\textbackslash nAnswer: " \\
& "Identify the single best answer. Your reply must consist solely of the letter corresponding to the selected choice (A, B, C, or D). No other text or commentary should be included.\textbackslash n\{sample\}\textbackslash nAnswer: " \\

\bottomrule

\end{tabularx}
\end{table*}

\begin{table*}[!tbh]
\centering
\small
\caption{Prompt formats for the CommonsenseQA dataset.}
\label{tab:prompt-format-cqa}
\begin{tabularx}{\textwidth}{@{}m{0.02cm}X@{}}
\toprule
 & \textbf{Prompt Text} \\ \midrule
\multirow{10}{*}{\rotatebox{90}{\begin{footnotesize}\textbf{Minimal Prompt Format}\end{footnotesize}}}   & "Here is a question with several possible answers: \{sample\}\textbackslash n"\\
& "Review the choices below and determine the best fit: \{sample\}\textbackslash n" \\
& "Could you provide a response to the following question:\textbackslash n\{sample\}\textbackslash nAnswer: " \\
& "Please answer the following question:\textbackslash n\{sample\}\textbackslash nAnswer: " \\
& "\{sample\}\textbackslash nI think it is " \\
& "\{sample\}\textbackslash nIt is " \\
& "\{sample\}\textbackslash nConsidering everything, I think it is " \\
& "\{sample\}\textbackslash nThe best answer seems to be " \\
& "\{sample\}\textbackslash nEverything points to " \\
& "\{sample\}\textbackslash n" \\ \midrule  

\multirow{19}{*}{\rotatebox{90}{\begin{footnotesize}\textbf{Instruction Prompt Format}\end{footnotesize}}}

& "Select the best answer from the given options. Respond only with the letter corresponding to the correct choice.\textbackslash n\{sample\}\textbackslash nAnswer: " \\
& "Choose the most appropriate answer among the given options. Return only the letter of the selected choice, without any explanation or additional text.\textbackslash n\{sample\}\textbackslash nAnswer: " \\
& "From the listed answer choices, select the most accurate one. Reply only with the letter of your selection (A, B, C, D, or E). Do not include any extra commentary or reasoning.\textbackslash n\{sample\}\textbackslash nAnswer: " \\
& "Identify the correct answer from the provided options. Your response must consist of a single uppercase letter (A, B, C, D, or E) matching your chosen answer. No explanations are needed.\textbackslash n\{sample\}\textbackslash nAnswer: " \\
& "Pick the best option and reply with only the corresponding letter (A, B, C, D, or E).\textbackslash n\{sample\}\textbackslash nAnswer: " \\
& "After reading the question and the provided answer choices, choose the single best answer. Your reply must be only the letter (A, B, C, D, or E) associated with the correct choice.\textbackslash n\{sample\}\textbackslash nAnswer: " \\
& "Treat this like a multiple-choice exam. Choose the correct answer and reply with only the letter of the option you select (A, B, C, D, or E). No explanations are required.\textbackslash n\{sample\}\textbackslash nAnswer: " \\
& "Select the correct option. Your entire response must consist of a single capital letter corresponding to your choice (A, B, C, D, or E) — no extra words, punctuation, or reasoning.\textbackslash n\{sample\}\textbackslash nAnswer: " \\
& "Pick the answer you think is best. Just send back the letter (like "A", "B", "C", "D" or "E") — no need to explain why.\textbackslash n\{sample\}\textbackslash nAnswer: " \\
& "Identify the single best answer. Your reply must consist solely of the letter corresponding to the selected choice (A, B, C, D, or E). No other text or commentary should be included.\textbackslash n\{sample\}\textbackslash nAnswer: " \\

\bottomrule

\end{tabularx}
\end{table*}

\begin{table*}[!tbh]
\centering
\small
\caption{Prompt formats for generating new prompt templates. The parts included in "\{\}" are replaced by the relevant data, such as task type, labels for the dataset or the examples of existing prompts.}
\label{tab:generation-prompt}
\begin{tabularx}{\textwidth}{@{}lX@{}}
\toprule
\textbf{Prompt Format} & \textbf{Prompt Text} \\ \midrule
 Minimal & I have a \{sentiment classification | news classification dataset | emotion detection | multi-choice question answering\} dataset. Can you give me some instructions that I could give to LLM so that it \{classifies | answers\} it? \textbackslash nSomething similar to: \{existing\_prompts\}\textbackslash nMake sure the prompt contains no instructions, is slightly leading but one could expect the LLM to answer something relevant. \\ \midrule

 Instruction & I have a \{sentiment classification | news classification dataset | emotion detection | multi-choice question answering\} dataset. Can you give me some instructions that I could give to LLM so that it \{classifies | answers\} it? \textbackslash nSomething similar to: \{existing\_prompts\}\textbackslash nMake sure it includes task instructions and possible labels that are \{labels\}. \\

\bottomrule

\end{tabularx}
\end{table*}

\textbf{Generation Performance Evaluation} When evaluating the generation performance, we use a simple string and regex matching to determine whether the generated text contains the given label. For the SST2 and AG News dataset, we first transform the text to lowercase and then search for each possible label. To determine whether it is contained in the text, we look for it as a separate word. If multiple labels are matched (e.g., text contains both "negative" and "positive" words for sentiment classification) we classify it as a failure case and assign label of -1. For the MCQA datasets (MMLU, ARC and CommonsenseQA), we continue the same way, but without converting the texts to lowercase, and instead of searching for uppercase possibilities (e.g., labels "A", "B", "C", "D", or "E") as separate words.

\section{Use of Generative Models}

During the writing of this paper, we have utilised the large language models in order to improve the grammar, sentence structure and the overall flow of some sections. In all cases, the texts were first written by the authors, passed to the LLM for improvement and carefully checked afterwards in order to check the meaning of the sentence has not changed.

\end{document}